\documentclass[lettersize,journal]{IEEEtran}

\def\ifundefined{\@ifundefined}

\usepackage{amsmath,amsfonts,amscd,amssymb}
\usepackage{cancel}
\usepackage{epsfig}
\usepackage{dsfont}
\usepackage{graphicx}
\usepackage{amsthm}
\usepackage[normalem]{ulem}
\usepackage{subfigure}
\usepackage{color}
\usepackage{hyperref}
\usepackage{bm}
\usepackage{lineno}
\usepackage{multirow}
\usepackage{amsmath}
\usepackage{cite}
\usepackage{lettrine}
\usepackage{diagbox} %
\usepackage{rotating}
\usepackage[super]{nth}
\usepackage{upgreek}
\usepackage{booktabs} %

\usepackage{makeidx}         %
\usepackage{multicol}        %
\usepackage[bottom]{footmisc}%
\usepackage{latexsym}
\usepackage{overpic}  
\usepackage{epstopdf}
\usepackage{rotating,color}

\usepackage{bbm}
\usepackage{algorithm,algpseudocode}
\usepackage{tikz}
\usepackage{makecell}
\usepackage[super]{nth}
\usepackage{euscript}
\usepackage{mathtools}

\usepackage{subeqnar}
\usepackage{textcomp,xcolor}
\definecolor{MatlabCellColour}{RGB}{250,250,250}
\definecolor{MatPurp}{rgb}{.625,.1406,.9375}

\DeclareMathAlphabet{\mathpzc}{OT1}{pzc}{m}{it}

\makeatletter
\let\start@align@nopar\start@align
\let\start@gather@nopar\start@gather
\let\start@multline@nopar\start@multline
\long\def\start@align{\par\start@align@nopar}
\long\def\start@gather{\par\start@gather@nopar}
\long\def\start@multline{\par\start@multline@nopar}
\makeatother

\newcommand{\be}{\begin{align}}
\newcommand{\ee}{\end{align}}
\newcommand{\bea}{\begin{eqnarray}}
\newcommand{\eea}{\end{eqnarray}}
\def\ea{\end{array}\right]}

\newcommand{\bfi}{\begin{figure}}
\newcommand{\efi}{\end{figure}}

\newcommand{\R}{\mathbb{R}}

\newcommand{\X}{\mathcal{X}}

\newcommand{\x}{\mathbf{x}}

\newcommand{\cmmnt}[1]{}

\newcommand{\inner}[1]{\langle {#1} \rangle}

\providecommand{\norm}[1]{\lVert#1\rVert}

\newtheorem{thm}{Theorem}  %

\newtheorem{defn}{Definition}       %

\begin{document}
\title{Kernel Operator-Theoretic Bayesian Filter for Nonlinear Dynamical Systems
	\thanks{This work was supported by the ONR grant N00014-23-1-2084.}
	\thanks{The authors are with the Computational NeuroEngineering Laboratory, University of Florida, Gainesville, FL 32611 USA  (e-mail: likan@ufl.edu; principe@cnel.ufl.edu).}}
\author{Kan Li, \IEEEmembership{Member,~IEEE} and Jos\'{e} C. Pr\'{i}ncipe, \IEEEmembership{Life Fellow,~IEEE}}

\maketitle

\begin{abstract}
Motivated by the surge of interest in Koopman operator theory, we propose a machine-learning alternative based on a functional Bayesian perspective for operator-theoretic modeling of unknown, data-driven, nonlinear dynamical systems. This formulation is directly done in an infinite-dimensional space of linear operators or Hilbert space with universal approximation property. The theory of reproducing kernel Hilbert space (RKHS) allows the lifting of nonlinear dynamics to a potentially infinite-dimensional space via linear embeddings, where a general nonlinear function is represented as a set of linear functions or operators in the functional space. This allows us to apply classical linear Bayesian methods such as the Kalman filter directly in the Hilbert space, yielding nonlinear solutions in the original input space. This kernel perspective on the Koopman operator offers two compelling advantages. First, the Hilbert space can be constructed deterministically, agnostic to the nonlinear dynamics. The Gaussian kernel is universal, approximating uniformly an arbitrary continuous target function over any compact domain. Second, Bayesian filter is an adaptive, linear minimum-variance algorithm, allowing the system to update the Koopman operator and continuously track the changes across an extended period of time, ideally suited for modern data-driven applications such as real-time machine learning using streaming data. In this paper, we present several practical implementations to obtain a finite-dimensional approximation of the functional Bayesian filter (FBF). Due to the rapid decay of the Gaussian kernel, excellent approximation is obtained with a small dimension. We demonstrate that this practical approach can obtain accurate results and outperform finite-dimensional Koopman decomposition.

\end{abstract}

\section{Introduction}
The recent surge of interest in the field of dynamical systems is closely tied to the emergence of \textit{statistical learning} and \textit{machine learning} (ML), fueled by the increasing abundance of data with advances in sensing and data acquisition technologies across all disciplines, especially engineering, biological, physical, and social sciences, revolutionizing our ability to analyze complex systems and relationships in real time, utilizing rich multi-modal and multi-fidelity time-series data.

A plethora of mathematical models and algorithms have arisen to meet this demand. Among them, deep neural networks (DNNs) stands out as particularly popular, with its foundational learning components inspired by biological neurons \cite{DL2015}. Algorithms grounded in DNNs have made significant strides in fields such as image classification, speech recognition, and natural language processing, surpassing even human-level performances \cite{SpeechDL2012, ImageNetNIPS2012, GANsNIPS2014, NLPsurvey2021}. While accuracy and interpretability are not necessarily mutually exclusive, in the field of \textit{deep learning}, the extraordinary achievements in performance often comes at the detriment of the other \cite{interDL2017}. Furthermore, these successes predominantly involve static pattern recognition or generation tasks. Deep learning methodologies become brittle and face significant challenges in dynamically changing environments \cite{DLbrittle2019, AIdynamicEnv2022}, such as those found in autonomous driving, due to their limited adaptation to temporal dynamics. In contrast, the dynamical system framework naturally accommodates translations in time, and the understanding of dynamics is of paramount importance.

Dynamical systems theory has a rich history of utilizing data to enhance modeling insights, develop parsimonious and interpretable models, and enable forecasting capabilities. In 1960, Kalman famously introduced a mathematical framework allowing the combination of observations and models through data assimilation techniques, particularly useful for forecasting and control \cite{Kalman60}. The integration of streaming data and dynamical models has a history of nearly seven decades. In the modern era, there is a growing trend to construct dynamical models directly from data using machine learning, especially in complex systems lacking first-principles models or where the correct state-space variable is unknown \cite{Schmidt2009, RAISSI2019686, LEE2020108973}. %

Nonlinearity presents a major challenge in the study of dynamical systems, giving rise to a wide array of phenomena such as bifurcations and chaos, which are observed across various disciplines. Despite its importance, there is no comprehensive mathematical framework available for the explicit and general characterization of nonlinear systems: the principle of linear superposition does not apply to nonlinear dynamical systems, resulting in a range of intriguing phenomena, such as harmonics and frequency shifts \cite{ModernKoopman2022}. Conversely, linear systems are fully characterized by their spectral decomposition (eigenvalues and eigenvectors), enabling the development of generic and computationally efficient algorithms for prediction, estimation, and control.

The past two decades has seen a renaissance of the Koopman operator theory \cite{Koopman31} due to its strong connections to data-driven modeling, offering a linear framework for the analysis and prediction of nonlinear behavior. Instead of directly analyzing the state space, it transforms a nonlinear dynamical system into a linear one in an infinite-dimensional space of observables (functions of the state). The Koopman operator advances these observables in time, offering a linear perspective on the evolution of the system, even if the underlying system is nonlinear, with its spectral decomposition fully characterizing the behavior of the nonlinear system. 

This operator theoretic perspective of dynamical systems was initially introduced to describe the evolution of measurements (flows) of Hamiltonian systems in 1931 \cite{Koopman31}. In the following year, it was generalized by Koopman and von Neumann to systems with a continuous eigenvalue spectrum \cite{Koopman32}. The Koopman operator also played a central role to the ergodic theory by von Neumann \cite{neumann1932pnas} and Birkhoff \cite{birkhoff1931pnas}. More recently, Rowley et al. \cite{rowley2009jfm} connected the Koopman mode decomposition, introduced by Mezi\'{c} in 2005 \cite{mezic2005nd}, with the dynamic mode decomposition (DMD), Schmid's numerical algorithm for fluid mechanics \cite{schmid2009dynamic}, based on the discrete Fourier transform (DFT) and the singular value decomposition (SVD), both of which provide unitary coordinate transformations \cite{brunton2019data}. Discovering tractable finite-dimensional representations of the Koopman operator is closely tied to identifying effective coordinate transformations that linearize the nonlinear dynamics. Judiciously chosen observables result in spatiotemporal features of the complex system that are physically meaningful and facilitate the application of manifold learning methods.

Koopman theory is connected to several related concepts. The Karhunen-Lo\`eve theorem \cite{loeve1960probability}, also known as the \textit{principal component analysis} (PCA) in statistics which generally traces back more than a century to Pearson \cite{Pearson1901}, provides a way to decompose a stochastic process into orthogonal functions, which can be seen as a form of optimal basis for representing the process. The relationship between the Karhunen–Lo\`eve theorem and Koopman operator theory lies in their use of eigenfunctions and spectral decomposition, but with different goals and contexts. Karhunen–Lo\`ve decomposes a stochastic process using eigenfunctions of the covariance operator, identifying dominant modes of variation, while Koopman decomposes the evolution of observables using eigenfunctions of the Koopman operator, revealing the underlying structure of a nonlinear dynamical system. Karhunen–Lo\`eve provides a linear representation of the data variability, useful for dimensionality reduction, while Koopman provides a linear representation of a nonlinear dynamical system's evolution in an infinite-dimensional space, aiding in the understanding and prediction of complex dynamics. Both methods can be used to analyze dynamical systems, but Koopman is specifically designed to handle nonlinear dynamics by leveraging linear operator theory on observables. Karhunen–Lo\`eve, on the other hand, focuses on the statistical properties of the data generated by the system. DMD, in the context of \textit{Koopman spectral analysis}, generates a collection of modes, each linked to a constant oscillation frequency and a decay/growth rate. Unlike PCA, which yields orthogonal modes without predetermined temporal characteristics, DMD captures inherent temporal patterns within each mode. It does not assume linearity, and, unlike \textit{proper orthogonal decomposition} (POD), the basis functions represented by the Koopman modes are not necessarily orthogonal \cite{DMDvsOPD2015}, with the growth rate determined by its real part, and its frequency identified by the imaginary part. Although DMD-based representations may be less parsimonious compared to PCA due to their non-orthogonal nature, they offer greater physical significance since each mode corresponds to a damped or driven sinusoidal behavior over time.

\begin{figure}[t]
	\centering
	\includegraphics[width=0.36\textwidth]{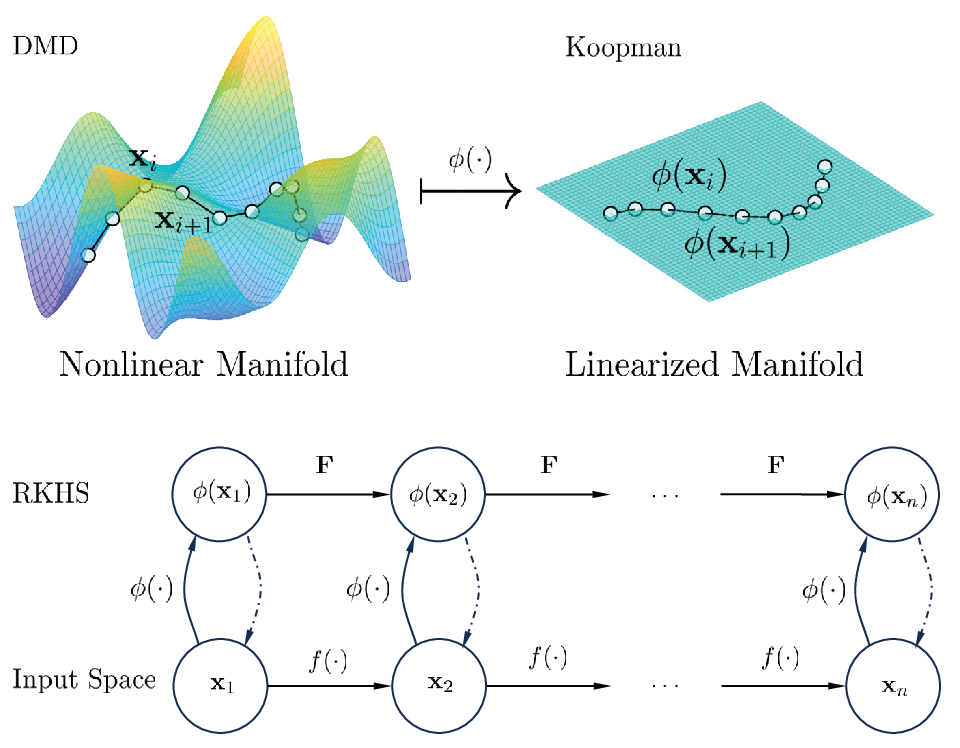}
	\caption{In a discrete-time nonlinear dynamical system, the nonlinearity generates a nonlinear manifold on which the time-series $\textbf{x}_i$ resides. DMD approximates the evolution on a nonlinear manifold using a least-squares fit linear dynamical system. The Koopman operator linearizes the space the data is embedded. Similarly, in the theory of the RKHS, the input data is mapped to a higher-dimensional functional space where a linear operator $\textbf{F}$ advances the states of the system, approximating the general nonlinear transition function $f(\cdot)$ in the input space.}
	\label{fig:manifold}
\end{figure}

Koopman operator theory is also closely related to the theory of \textit{reproducing kernel Hilbert space} (RKHS) \cite{RKHS1950} or kernel methods \cite{kernel2008}. Both operate in a potentially infinite feature space and are data-driven. Koopman operates in the space of observables, which can be infinite-dimensional, while kernel methods operate in a functional space induced by a reproducing kernel, which can be infinite-dimensional depending on the kernel choice. Koopman linearizes the dynamics by considering the temporal evolution in an infinite-dimensional space of observables. Kernel methods, on the other hand, simplify the design of classification or regression problems by mapping data into a higher-dimensional (potentially infinite) linear feature space (nonlinear relative to the input space) where linear methods can be applied. The effectiveness of Koopman methods hinges primarily on selecting an appropriate set of observables. An important practical consideration is the computational cost as the number of observables or feature dimension increases. Central to the idea of kernel methods is the \textit{kernel trick}: instead of considering the actual space of features, a kernel function is used to consider a large class of potential observables without considering the actual observation vector, making the computation efficient and tractable. Using this compact representation of the feature space, kernelized versions of DMD implicitly compute the inner products required for the Koopman decomposition \cite{kDMD2015,FUJII201994}. Fig. \ref{fig:manifold} illustrates the underlying principle in the RKHS and Koopman operator theoretic view.

Despite the overwhelming usefulness of both methods, there still exist significant gaps between the two theories. Koopman operator theory does not provide insights on how to construct the space of infinite-dimensional observables or its finite approximation for a particular application, and its performance is sensitive to the choice of observables. Physically interpretable spatiotemporal features have to be judiciously chosen and can be prohibitively challenging and may rival that of first-principles modeling.  Conventional kernel methods, on the other hand, take the infinite-dimensional space for granted (automatically induced by an appropriate reproducing kernel function), but lack instructions on how to properly handle dynamics in the RKHS. 

There are also many practical issues using kernels. Despite circumventing the need to consider the actual dimension of the RKHS, the computational complexity of kernel methods can grow linearly and superlinearly with the number of data points or time steps, without introducing additional sparsification techniques. This becomes especially problematic for time series with complex dynamics and where the temporal dimension is significantly larger than the spatial dimension, e.g., streaming data. Furthermore, not having an explicit mapping of the input data gives rise to the problem of finding the \textit{pre-image} of a feature vector in the RKHS induced by a kernel.

The major contribution of this paper is a novel algorithm that bridges the two concepts for dynamical systems analysis by combining the best of both worlds: a completely data-driven approach that maps nonlinear dynamics to a finite-dimensional RKHS using predetermined, explicit mapping, and applies classical linear recursive Bayesian estimation on the feature-space state variables to derive a nonlinear solution in the original input space.

The current development represents the next phase in the evolution of research on kernel adaptive filtering (KAF) for discrete-time nonlinear dynamical systems. We first proposed a full state-space representation of a general nonlinear dynamical model over an RKHS and its learning algorithm called the \textit{kernel adaptive autoregressive-moving-average} (KAARMA) in \cite{KAARMA}, then followed up with a general nonlinear Bayesian inference for high-dimensional state estimation or the \textit{functional Bayesian filter} (FBF) \cite{FBF2022}. In \cite{NoTrick2019}, we showed that an equivalent, finite-dimensional Hilbert space can be constructed using numerical integration methods such as Gaussian quadrature (GQ) and Taylor series (TS) expansion that leads to significant savings in performing KAF on streaming data with comparable performance to their infinite dimensional counterparts and avoids the pre-image problem plaguing conventional kernel methods. We combine these features here to propose an explicit-space functional Bayesian filter (expFBF) that constructs a suitable finite-dimensional RKHS \textit{a priori} that linearizes the dynamics, then recursively learn and track the linear dynamics using Bayesian update.

The Kalman filter is the most popular Bayesian method that provides an analytical linear solution to estimate the target state while minimizing the error associated with the estimation. It is optimal for linear dynamical systems under additive noise with finite second-order moments, recursively estimating the state by combining predictions from a model with measurements, updating the state estimates and uncertainties over time. The major shortcoming of the Kalman approach is that the dynamics are assumed known. Kernel methods can map the input data into a higher-dimensional feature space where the dynamics are linearly separable, allowing the use of Kalman filtering in that space, but without precise knowledge of  the system dynamics (real-world applications often involve unknown and nonstationary transformations), the performance suffers due to parametric errors in the model. The FBF \cite{FBF2022} is a purely data-drive approach that constructs a state-space representation of the data in an RKHS and jointly estimates the model parameters and the states of the linear dynamical system, recursively updating not only the weight changes, but also the state variables (activities in the learning network) in response to the weights while minimizing their respective uncertainties. This enables continuous learning and tracking of complex dynamics over a long-term that the conventional Koopman operator theory cannot. We summarize the key attributes of both paradigms in Table \ref{tab:Attributes}. 

\begin{table}[!t]
	\begin{center}
	\setlength{\tabcolsep}{6pt}
	\scriptsize
	\caption{Key features of the Koopman theory and RKHS.}
	\label{tab:Attributes}
	\begin{tabular}{|l|c|c|}
		\hline		
		\diagbox[width=8em]{Property}{Theory} &  \thead{\textbf{Koopman Operator} \\ (\textbf{DMD})} &  \thead{\textbf{KAF} \\(\textbf{Explicit-Space FBF})}\\ \hline
		\multirow{ 2}{*}{\textbf{Feature Space}} & unknown & induced by kernel \\ 
		& (requires judicious selection) & (predetermined) \\ \hline
		\multirow{ 2}{*}{\textbf{Dimension}} & \multirow{ 2}{*}{finite to $\infty$} & finite to $\infty$ \\
		& & (finite) \\ \hline
		\textbf{Data} & data-driven + expert knowledge & data-driven \\ 
		& (batch) & (online) \\ \hline
		\textbf{Duration} & short-term & long-term \\ \hline
		\textbf{Dynamics} & nonlinear & nonlinear \\ \hline
		\multirow{ 2}{*}{\textbf{Stationarity}} & \multirow{ 2}{*}{stationary} & nonstationary \\
		& & (tracks the posterior) \\ \hline
		\multirow{ 2}{*}{\textbf{Performance}} & sensitive to choice & robust \\
		& of observables & (minimum variance) \\ \hline
	\end{tabular}
	\end{center}
	\normalsize
\end{table}

Motivated by the above discussion, we propose an operator-theoretic data-fusion approach for learning the Koopman operator, formulated directly in the finite-dimensional space of linear operators or RKHS. The rest of this paper is organized as follows. In Section \ref{Sec:Bayesian}, state-space model in the RKHS is reviewed, followed by the finite-dimensional explicit-space RKHS. In Section \ref{Sec:expFBF} we present the proposed explicit-space functional Bayesian filter algorithm. Section \ref{Sec:Results} shows the experimental results for unknown chaotic time series and nonlinear partial differential equation (PDE), comparing our novel method with several existing algorithms including the DMD and Koopman methods. Finally, Section \ref{Sec:Conclusion} concludes this paper.
 
\section{Bayesian filtering in the RKHS} \label{Sec:Bayesian}
\begin{figure}[h]
	\centering
	\includegraphics[width=0.4\textwidth]{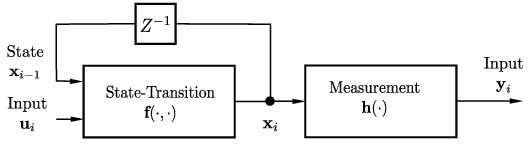}
	\caption{General state-space model for dynamical system.}
	\label{fig:model}
\end{figure}

For nonlinear dynamical systems, we are interested in estimating the state variables recursively via the sequence of noisy measurements (observations) dependent on the state. Let a dynamical system (Fig. \ref{fig:model}) be defined in terms of a general nonlinear state-transition and observation functions, $\textbf{f}(\cdot,\cdot)$ and $\textbf{h}(\cdot)$, respectively, 
\begin{align}
	\textbf{x}_{i} &= \textbf{f}(\textbf{x}_{i-1},\textbf{u}_{i})+\bm{w}_{i-1}\label{eq:ssm1}\\
	\textbf{y}_i &= \textbf{h}(\textbf{x}_{i})+\bm{v}_{i}\label{eq:ssm2}
\end{align}
where 
\begin{align}
	\textbf{f}(\textbf{x}_{i-1},\textbf{u}_i) &\stackrel{\Delta}{=} \left[f^{(1)}(\textbf{x}_{i-1},\textbf{u}_i),\cdots,f^{(n_x)}(\textbf{x}_{i-1},\textbf{u}_i)\right]^T\\
	\textbf{h}(\textbf{x}_i) &\stackrel{\Delta}{=} \left[h^{(1)}(\textbf{x}_i),\cdots,h^{(n_y)}(\textbf{x}_i)\right]^T
\end{align}
with input $\textbf{u}_i\in\mathbb{R}^{n_u}$, state $\textbf{x}_i\in\mathbb{R}^{n_x}$, output $\textbf{y}_i\in\mathbb{R}^{n_y}$, additive dynamic noise $\bm{w}_i$, and observation noise $\bm{v}$ are statistically independent processes of zero mean and known covariance matrices, and the parenthesized superscript $^{(k)}$ denotes the $k$-th column of a matrix or the $k$-th component of a vector. For versatility, the input, state, and output variables have independent degrees of freedom (dimensionality).

We can simplify the expression above by expressing the dynamical system equations (\ref{eq:ssm1}-\ref{eq:ssm2}) in terms of a new hidden state vector
\begin{align}
	\textbf{s}_{i} &\stackrel{\Delta}{=}
	\begin{bmatrix}
		\textbf{x}_{i} \\[0.3em]
		\textbf{y}_i\\[0.3em]
	\end{bmatrix} = \begin{bmatrix}
		\textbf{f}(\textbf{x}_{i-1},\textbf{u}_{i})\\[0.3em]
		\textbf{h}\circ\textbf{f}(\textbf{x}_{i-1},\textbf{u}_{i})\\[0.3em]
	\end{bmatrix} \label{eq:aug_state}\\
	\textbf{y}_i &= \textbf{s}_{i}^{(n_s-n_y+1:n_s)} =  \underbrace{\begin{bmatrix}
			\textbf{0} & \textbf{I}_{n_y} \\[0.3em]
	\end{bmatrix}}_{\mathds{I}}\begin{bmatrix}
		\textbf{x}_{i} \\[0.3em]
		\textbf{y}_i\\[0.3em]
	\end{bmatrix}\label{eq:selector}
\end{align}
where $\textbf{I}_{n_y} $ is an $n_y\times n_y$ identity matrix, \textbf{0} is an $n_y\times n_x$ zero matrix, and $\circ$ is the function composition operator. This augmented state vector $\textbf{s}_i\in\mathbb{R}^{n_s}$ is formed by concatenating the output $\textbf{y}_i$ with the original state vector $\textbf{x}_i$, i.e., an augmented state with dimension $n_s = n_x+n_y$. With this rewriting, the measurement equation simplifies to a fixed selector matrix $\mathds{I}\stackrel{\Delta}{=} \begin{bmatrix}
	\textbf{0} & \textbf{I}_{n_y} \\[0.3em]
\end{bmatrix}$. Despite the parsimonious structure of \eqref{eq:selector}, there is no restriction on the measurement equation, as $\textbf{h}\circ\textbf{f}$ in \eqref{eq:aug_state} is its own set of general nonlinear equations, i.e., this is functionally equivalent to the generative model in Fig. \ref{fig:model}.

Next, we define an equivalent transition function $\textbf{g}(\textbf{s}_{i-1},\textbf{u}_{i})=\textbf{f}(\textbf{x}_{i-1},\textbf{u}_{i})$ using the augmented state variable $\textbf{s}$ as argument. The dynamic system in (\ref{eq:ssm1}-\ref{eq:ssm2}) becomes
\begin{align}
	\textbf{x}_{i} &= \textbf{g}(\textbf{s}_{i-1},\textbf{u}_{i})\label{eq:nssm1}\\
	\textbf{y}_i &= \textbf{h}(\textbf{x}_{i}) = \textbf{h}\circ\textbf{g}(\textbf{s}_{i-1},\textbf{u}_{i}).\label{eq:nssm2}
\end{align}

The theory of RKHS enables the lifting of the dynamics to a potentially infinite-dimensional space via linear embeddings, where a general nonlinear function is represented as a set of linear weights (operators) $\bm{\Omega}$ in the feature (functional) space. To learn the general continuous nonlinear transition and observation functions, $\textbf{g}(\cdot,\cdot)$ and $\textbf{h}\circ \textbf{g}(\cdot,\cdot)$, respectively, we map the augmented state vector $\textbf{s}_{i}$ and the input vector $\textbf{u}_{i}$ into separate RKHSs as $\varphi(\textbf{s}_{i})\in\mathcal{H}_s$ and $\phi(\textbf{u}_{i})\in\mathcal{H}_u$.  By the \textit{representer theorem}, the state-space model defined by (\ref{eq:nssm1}-\ref{eq:nssm2}) can be expressed as the following set of weights (functions) in the joint RKHS  $\mathcal{H}_{su}\stackrel{\Delta}{=}\mathcal{H}_{s}\otimes\mathcal{H}_{u}$: 
\begin{align}
	{\bm\Omega}\stackrel{\Delta}{=}{\bm\Omega}_{\mathcal{H}_{su}}\stackrel{\Delta}{=} \begin{bmatrix}
		\textbf{g}(\cdot,\cdot)\\[0.3em]
		\textbf{h}\circ \textbf{g}(\cdot,\cdot) \\[0.3em]\end{bmatrix}
\end{align}
where $\otimes$ is the tensor-product operator. This formulation preserves the functionalities of the separate state-transition and observation equations while consolidating them into a single set of weights in the RKHS (i.e., a single, parsimonious network), which greatly facilitates the construction of an empirical model and the adaptation of parameters.

Finally, the state-space model in the RKHS becomes
\begin{align}
	\textbf{s}_{i}&={\bm\Omega}^T\psi(\textbf{s}_{i-1},\textbf{u}_{i})\label{algGenerateState}\\
	\textbf{y}_i&=\mathds{I}\textbf{s}_i\label{algOutput}
\end{align}
where new features were defined using a tensor-product in \cite{KAARMA} and \cite{FBF2022} as
\begin{align} \psi(\textbf{s}_{i-1},\textbf{u}_{i})\stackrel{\Delta}{=}\varphi(\textbf{s}_{i-1})\otimes\phi(\textbf{u}_i)\in \mathcal{H}_{su}.
\end{align}

\subsection{Finite-Dimensional RKHS}
In \cite{NoTrick2019}, we demonstrated that for continuous shift-invariant kernel functions, such as the Gaussian, one can construct finite-dimensional explicit mappings or deterministic features that define an approximately equivalent reproducing kernel and achieve similar performances in KAF using a fixed finite-dimensional weight vector. This new kernel can approximate the original universal kernel to arbitrary accuracy with appropriate truncation of the Taylor series or Hermite polynomial expansion. This not only enables efficient implementations of KAF, but also, more importantly, allows access to the pre-image or reverse mapping, without compromising performance. For completeness, we summarize the findings below.

\begin{thm}[Bochner, 1932\cite{Bochner1959}]
	A continuous shift-invariant properly-scaled kernel $k(\x,\x\prime)=k(\x-\x\prime):\R^d\times\R^d\rightarrow\R$, and $k(\x,\x)=1,\forall\x$, is positive definite if and only if $k$ is the Fourier transform of a proper probability distribution.
\end{thm}
The corresponding kernel can be expressed in terms of its Fourier transform $p(\bm{\omega})$ (a probability distribution or pdf) as
\begin{align}
	k(\x-\x\prime)&=\int_{\R^d}p(\bm{\omega})e^{j\bm{\omega}^\intercal(\x-\x\prime)}d\bm{\omega}\label{eq:kintergral}\\
	&={\rm E}_{\bm{\omega}}\left[e^{j\bm{\omega}^\intercal(\x-\x\prime)}\right]={\rm E}_{\bm{\omega}}\left[\langle e^{j\bm{\omega}^\intercal\x},e^{j\bm{\omega}^\intercal\x\prime}\rangle\right]\label{eq:Fourier_kernel}
\end{align}
where $\langle \cdot,\cdot \rangle$ is the Hermitian inner product $\langle \x,\x\prime\rangle =\sum_i\x_i\overline{\x\prime_i}$, and $\langle e^{j\bm{\omega}^\intercal\x},e^{j\bm{\omega}^\intercal\x\prime}\rangle$ is an unbiased estimate of the properly scaled shift-invariant kernel $k(\x-\x\prime)$ when $\bm{\omega}$ is drawn from the probability distribution $p(\bm{\omega})$. 

Typically, we ignore the imaginary part of the complex exponentials to obtain a real-valued mapping. This can be further generalized by considering the class of kernel functions with the following construction
\begin{align}
	k(\x,\x\prime)=\int_{\mathcal{V}}z(\mathbf{v},\x)z(\mathbf{v},\x\prime)p(\mathbf{v})d\mathbf{v}
\end{align}
where $z:\mathcal{V}\times\mathcal{X}\rightarrow\R$ is a continuous and bounded function with respect to $\mathbf{v}$ and $\x$. The kernel function can be approximated by its Monte-Carlo estimate
\begin{align}
	\tilde{k}(\x,\x\prime)=\frac{1}{D}\sum^{D}_{i=1}z(\mathbf{v}_i,\x)z(\mathbf{v}_i,\x\prime) \label{eq:approx_kernel}
\end{align} 
with feature space $\mathbb{R}^D$ and $\{\mathbf{v}_i\}^D_{i=1}$ sampled independently from the spectral measure.

Instead of using random features to construct the explicit Hilbert space, which yields large variances in performance, related to the location of the expansion points $v_i$, especially for smaller dimensions, we can approximate the integral in \eqref{eq:kintergral} with a discrete sum of judiciously selected
points. An explicit finite-dimensional RKHS using Gaussian quadrature can accurately approximate the continuous shift-invariant kernel including all sub-Gaussian densities as $k(\x,\x\prime) : \X \times \X \to \R \approx \widetilde{\phi}(\x)^\intercal \widetilde{\phi}(\x\prime)$, where $\widetilde{\phi} : \X \to \R^D$ defines an explicit mapping to a finite-dimensional feature space \cite{Dao2017, NoTrick2019}. Here, we are primarily concerned with the Gaussian kernel since it is universal, i.e., approximates an arbitrary continuous function uniformly on any compact subset of the input space \cite{UniversalKernels2006}. Nonetheless, any kernel function can be approximated by its convolution with the Gaussian, resulting in a significantly smaller approximation error than the noise from data generation.

Specifically, we consider the Gauss–Hermite quadrature, which employs Hermite polynomials which are the eigenfunctions of the Gaussian kernel and form a Hermite spectral decomposition. To extend one-dimensional GQ (accurate for polynomials up to degree $R$) to higher dimensions, the following  must be satisfied
\begin{align}
	\int_{\R^d} p(\bm{\omega}) \prod_{l=1}^d 
	(\mathbf{e}_l^\intercal\bm{\omega})^{r_l} d\bm{\omega} = \sum_{i=1}^D a_i \prod_{l=1}^d (\mathbf{e}_l^\intercal \bm{\omega}_i)^{r_l}\label{eqnPolyExactConstraints}
\end{align}
for all $r \in \mathbb{N}^d$ such that $\sum_l r_l \leq R$, where $\mathbf{e}_l$ are the standard basis vectors.

To construct GQ features, we can randomly select points $\bm{\omega}_i$, then solve \eqref{eqnPolyExactConstraints} for the weights $a_i$ using non-negative least squares (NNLS) algorithm. This works well in low dimensions, but for larger values of d and R, grid-based quadrature rules can be constructed more efficiently.

The major drawback of the grid-based construction is the lack of fine tuning for the feature dimension. Since the number of samples extracted in the feature map is determined by the degree of polynomial exactness, even a small incremental change can cause a significant increase in the number of features. \emph{Subsampling} according to the distribution determined by their weights is used to combat both the curse of dimensionality and the lack of detailed control over the exact sample number. There are also data-adaptive methods to choose a quadrature rule for a predefined number of samples \cite{Dao2017}.

Another useful finite-dimensional approximation is the Taylor series expansion polynomial features \cite{Cotter2011}, a special case in the class of analytic positive definite multivariate functions called the power series kernels \cite{Zwicknagl2009}, where each term is expressed as a sum of matching monomials in $\x$ and $\x\prime$, i.e.,
\begin{align}\label{eq:express_Gaussian}
	k(\x, \x\prime) = e^{-\frac{\|\x- \x\prime\|^2}{2\sigma^2}} =
	e^{-\frac{\|\x\|^2}{2\sigma^2}} e^{\frac{\langle \x, \x\prime\rangle}{\sigma^2}} e^{-\frac{\|\x\prime\|^2}{2\sigma^2}}.
\end{align}
We can easily factor out the terms that depend on $\x$ and $\x\prime$ independently. The cross term in \eqref{eq:express_Gaussian} can be expressed as a power series or infinite sum using Taylor polynomials as
\begin{align}\label{eq:scalar_taylor}
	e^{\frac{\inner{\x,\x\prime}}{\sigma^2}} = \sum_{n=0}^{\infty}
	\frac{1}{n!}\left(\frac{\inner{\x,\x\prime}}{\sigma^2}\right)^n.
\end{align}
This produces the following explicit feature map
\begin{align}
	\phi\left( \x \right) = e^{-\frac{\|\x\|^2}{2\sigma^2}}\sum_{\alpha\in\mathbb{N}^d_0} \frac{1}{\sigma^{|\alpha|}\sqrt{\alpha!}}
	{\x}^\alpha\label{eq:taylor_features}
\end{align}
yielding 
\begin{align}
	k(\x,\x\prime) &= \inner{\phi(\x),\phi(\x\prime)} \\
	&= e^{-\frac{\|\x\|^2}{2\sigma^2}}\sum_{\alpha\in\mathbb{N}^d_0} \frac{1}{\sigma^{2|\alpha|}\alpha!}
	{\x}^\alpha {\x\prime}^\alpha e^{-\frac{\|\x\prime\|^2}{2\sigma^2}}.
\end{align}

We can truncate the infinite sum to derive an approximation that is exact up to polynomials of degree $r$. The dimensionality and coordinates (unique monomials) of the TS approximation can be computed efficiently using the multinomial theorem, where we have a total of $D = \binom{d+r}{r}$ features of degree up to $r$ and $\binom{d+n-1}{n}$ features of degree $n$, i.e.,
\begin{align}
	\label{eq:taylor_kernel_expansion} \tilde{k}(\x,\x\prime)&=\inner{\widetilde{\phi}(\x),\widetilde{\phi}(\x\prime)}\\\nonumber
	&= e^{-\frac{\|\x\|^2+\|\x\prime\|^2}{2\sigma^2}} \sum_{n=0}^{r} \frac{1}{n!}\left(\frac{\inner{\x,\x\prime}}{\sigma^2}\right)^n
\end{align}
where
\begin{align}\label{eq:multinomial_exp}
	\inner{\x,\x\prime}^n=\left(\sum_{i=1}^d \x_i \x\prime_i\right)^n=\sum_{|\alpha| = n} \binom{n}{\alpha}\x^\alpha {\x\prime}^\alpha
\end{align}
and $\alpha =(\alpha_1,\alpha_2,\cdots,\alpha_d)$ and $\x^\alpha = \x^{\alpha_1}_1\x^{\alpha_2}_2\cdots\x^{\alpha_d}_d$.

The Taylor series for $e^x$ has an infinite radius of convergence. The truncation error is bounded using Taylor's theorem:
\begin{align}
	\left|e^{\frac{\inner{\x,\x\prime}}{\sigma^2}} -\sum_{n=0}^{r}
	\frac{1}{n!}\left(\frac{\inner{\x,\x\prime}}{\sigma^2}\right)^n\right| \leq \left| 
	\frac{e^{\frac{\inner{\x,\x\prime}}{\sigma^2}}\left(\frac{\inner{\x,\x\prime}}{\sigma^2}\right)^{r+1}}{(r+1)!}\right|\label{eq:TaylorTheorem}
\end{align}
which yields the following upper bound for the difference in kernel evaluations
\begin{align}
	\left|k(\x,\x\prime)- \tilde{k}(\x,\x\prime)\right|\leq \frac{\left(\frac{\norm{\x}\norm{\x\prime}}{\sigma^2}\right)^{r+1}}{(r+1)!}
\end{align}
where $\left|\inner{\x,\x\prime}\right|\leq\norm{\x}\norm{\x\prime}$ by the Cauchy-Schwarz inequality. For simplicity, we will express the Gaussian kernel in terms of the kernel parameter $a=\frac{1}{2\sigma^2}$, i.e.,
$k(\x, \x\prime) = e^{-a\|\x- \x\prime\|^2}$.

\section{Functional Bayesian Filtering over Finite-Dimensional RKHS} \label{Sec:expFBF}
The Bayesian solution provides a unifying framework for solving the recursive estimation of state variables in a dynamical system. Since the functions are represented as linear operators or weights $\bm{\Omega}$ in the RKHS, we can apply Kalman filtering to the explicitly mapped points in the finite-dimensional Hilbert space. However, it cannot be applied directly due to the unknown dynamics. As shown in \cite{FBF2022}, the Bayesian framework further allows us to jointly estimate the states and the weights by treating $\bm{\Omega}$ as part of the state variables.

First, we construct the following linear form of the state-space representation:
\begin{align}
	{\mathbbm{s}}_i & = \textbf{F}_{i-1}{\mathbbm{s}}_{i-1} + \bm{w}_{i-1} \label{eq:StateTrans} \\
	{\textbf{y}}_i  & = \textbf{H}_i{\mathbbm{s}}_i + \bm{v}_{i}\label{eq:Meas}
\end{align}
where the super-augmented state vector is defined as
\begin{align}
	\mathbbm{s}_i \stackrel{\Delta}{=} \begin{bmatrix}
		                                   \textbf{s}_i  \\[0.3em]
		                                   \bm{\Omega}_i \\[0.3em]
	                                   \end{bmatrix}\label{Eq:AugState}
\end{align}
with $\textbf{s}_{i} =[\textbf{x}_{i},  \textbf{y}_i]^\intercal$, same as in \eqref{eq:aug_state}, and we treat the weight matrix $\bm{\Omega}_i$ in the RKHS at time $i$ as an $n_{\bm{\Omega}}$-dimensional (finite for explicit RKHS) vector rather than a matrix, via an orderly arrangement of the weight parameters (e.g., stacking each transposed row vertically). Next, the state transition matrix can be expressed in block form as
\begin{align}
	\textbf{F}_i = \begin{bmatrix}
		               \textbf{F}_1(i) & \textbf{F}_2(i)              \\[0.3em]
		               \textbf{0}      & \textbf{I}_{n_{\bm{\Omega}}} \\[0.3em]
	               \end{bmatrix} \label{eq:F}
\end{align}
where $\textbf{F}_1(i)$ is an $n_s\times n_s$ matrix, $\textbf{F}_2(i)$ is an $n_s\times n_{\bm{\Omega}}$ matrix, and $\textbf{I}_{n_{\bm{\Omega}}}$ is an $n_{\bm{\Omega}}\times n_{\bm{\Omega}}$ identity matrix. The unknown weight vector is assumed to be invariant, i.e., $\bm{\Omega}_{i+1} = \bm{\Omega}_i$. State transition equation \eqref{eq:StateTrans} becomes
\begin{align}
	\begin{bmatrix}
		\textbf{s}_i  \\[0.3em]
		\bm{\Omega}_i \\[0.3em]
	\end{bmatrix} = \begin{bmatrix}
		                \textbf{F}_1(i) & \textbf{F}_2(i)              \\[0.3em]
		                \textbf{0}      & \textbf{I}_{n_{\bm{\Omega}}} \\[0.3em]
	                \end{bmatrix}\begin{bmatrix}
		                             \textbf{s}_{i-1} \\[0.3em]
		                             \bm{\Omega}_{i-1}    \\[0.3em]
	                             \end{bmatrix} + \bm{w}_{i-1}. \label{Eq:SSM}
\end{align}
Since the network output $\textbf{y}_i$ is a subvector of the hidden state $\textbf{s}_{i}$, the measurement function $\textbf{H}_i$ in \eqref{eq:Meas} becomes a simple projection onto the last $n_y$ components of $\textbf{s}_i$
\begin{align}
	\textbf{y}_i = \textbf{H} \begin{bmatrix}
		                          \textbf{s}_i  \\[0.3em]
		                          \bm{\Omega}_i \\[0.3em]
	                          \end{bmatrix} + \bm{v}_{i}
\text{, where }
	\textbf{H} \stackrel{\Delta}{=} \begin{bmatrix} \mathds{I} & \textbf{0} \end{bmatrix}\label{eq:H}
\end{align}
with $\textbf{H}\in \mathbb{R}^{n_y\times (n_s+n_{\bm{\Omega}})}$ and $\mathds{I}\stackrel{\Delta}{=} \begin{bmatrix}
		\textbf{0} & \textbf{I}_{n_y} \\[0.3em]
	\end{bmatrix}\in \mathbb{R}^{n_y\times n_s}$ being a fixed selector matrix.

In \cite{FBF2022}, we used the tensor product kernel to construct the infinite-dimensional joint RKHS for the state and control inputs $\mathcal{H}_{su}$. Finite-dimensional explicitly defined RKHS allows us a greater degree of flexibility on how to combine the two feature spaces. Without loss of generality, we use the canonical linear state-space representation to propagate the states with a sum instead of product, replacing \eqref{algGenerateState} with
\begin{align} 
\textbf{s}_{i} = \textbf{A}_{i}\widetilde{\psi}(\textbf{s}_{i-1}) + \textbf{B}_{i} \widetilde{\phi}(\textbf{u}_{i})\label{eq:exp_state_transition}
\end{align}
where $\textbf{A}_{i}\in \mathbb{R}^{n_s\times n_{\Omega_s}}$ and $\textbf{B}_{i}\in \mathbb{R}^{n_s\times n_{\Omega_u}}$ are the weight vectors for the state features $\widetilde{\psi}(\textbf{s})$ and input features $\widetilde{\phi}(\textbf{u})$, respectively. Note, the tensor-product construction is more expressive (explicitly defines all the cross terms between the state and control variables as features). However, this creates more complexity in the explicit space as the number of parameters becomes multiplicative, instead of additive. Furthermore, \eqref{eq:exp_state_transition} is a direct analog to the model used in discrete-time linear time-invariant (LIT) control system.

For applications involving known nonlinear dynamics or observable state variables, we can propagate the states directly in the RKHS, i.e.,
\begin{align} 
	\widetilde{\psi}(\textbf{s}_{i}) = \textbf{A}_{i}\widetilde{\psi}(\textbf{s}_{i-1}) + \textbf{B}_{i} \widetilde{\phi}(\textbf{u}_{i}).\label{eq:RKHS_state_transition}
\end{align}
For the case involving only state transitions, e.g., for many nonlinear PDE problems, \eqref{eq:RKHS_state_transition} simplifies to 
\begin{align} 
	\psi(\textbf{s})_{i} = \textbf{A}_{i}\psi(\textbf{s})_{i-1}.\label{eq:exp_only_state_transition}
\end{align}

Since the RKHS can be explicitly defined and is finite-dimensional by construction, to obtain the inverse map, we can simply augment the feature space by concatenating the original state variables $\textbf{s}_i$ with the mapped feature-space states (observables) $\psi(\textbf{s})_{i}$ and output the first $n_s$ components at each time step, i.e.,
\begin{align}
	\begin{bmatrix}
		\textbf{s}_{i}  \\[0.3em]
		\psi(\textbf{s})_{i} \\[0.3em]
	\end{bmatrix} = \textbf{A}_{i}
		\begin{bmatrix}
		\textbf{s}_{i-1}  \\[0.3em]
		\psi(\textbf{s})_{i-1} \\[0.3em]
	\end{bmatrix}.\label{eq:inverse_map_concat}
\end{align}
This is consistent with Koopman theory that allows a broader set of observables which often includes the original states as a subset. This way, the algorithm produces both a state estimation as well as the linearized nonlinearity. 	
\subsection{Bayesian Update} 
From the state-space model in \eqref{Eq:SSM}, the state-transition matrix block in \eqref{eq:F} consists of the following Jacobians
\begin{align}
	\textbf{F}_1(i) & = \frac{\partial \textbf{s}_i}{\partial \textbf{s}_{i-1}} \label{Eq:F1}
\end{align}
and
\begin{align}
	\textbf{F}_2(i) = \frac{\partial \textbf{s}_i}{\partial \bm{\Omega}_i}.
\end{align}

From \eqref{eq:RKHS_state_transition}, the Jacobian with respect to the weights can be written in block form as $\textbf{F}_2(i) = \begin{bmatrix}\textbf{F}^{(\textbf{A})}_2(i) & \textbf{F}^{(\textbf{B})}_2(i)\end{bmatrix}$
with
\begin{align}
	\textbf{F}^{(\textbf{A})}_2(i) = \frac{\partial \textbf{s}_i}{\partial \textbf{A}(i)}\label{Eq:F2A}
\end{align}
and
\begin{align}
	\textbf{F}^{(\textbf{B})}_2(i) = \frac{\partial \textbf{s}_i}{\partial \textbf{B}(i)}.\label{Eq:F2B}
\end{align}
Note here that the weight vector is arranged element-wise in a single column as
\begin{align}
	\bm{\Omega}\stackrel{\Delta}{=} \scalebox{.95}{$\begin{bmatrix}
		                                \bm{\alpha}_{11},
		                                \bm{\alpha}_{12},
		                                \cdots,
		                                \bm{\alpha}_{n_s n_{\widetilde{\psi}(s)}} |
		                                \bm{\beta}_{11},
		                                \bm{\beta}_{12},
		                                \cdots,               
		                                \bm{\beta}_{n_s n_{\widetilde{\phi}(u)}}
	                                \end{bmatrix}^\intercal$}\label{AB}
\end{align}
where $\bm{\alpha}_{ij}$ and $\bm{\beta}_{ij}$ are the elements of $\textbf{A}$ and $\textbf{B}$, respectively.

The \textit{a priori} estimate covariance matrix is given by
\begin{align}
	\textbf{P}_{i|i-1} = \textbf{F}_i\textbf{P}_{i-1|i-1}\textbf{F}_i^\intercal+\textbf{Q}_{i-1}\label{eq:CovEst}
\end{align}
where 
\begin{align}
	\textbf{Q}_i =\mathbf{E}\left[\bm{w}_i \bm{w}^\intercal_i\right] 
\end{align}
is the process noise covariance matrix. In general, $\textbf{Q}$ may have non-zero off-diagonal elements (indicating correlated state variables). However, for simplicity, we will assume diagonal matrices for the noise covariance matrices (unless specified otherwise) and, likewise, initialize the unknown state covariance $\textbf{P}$ as a diagonal matrix.

The block structure of the state transition matrix $\textbf{F}$ yields the following estimated state covariance matrix decomposition
\begin{align}
	\textbf{P} = \begin{bmatrix}
		\textbf{P}_1 & \textbf{P}_2 \\[0.3em]
		\textbf{P}_3 & \textbf{P}_4 \\[0.3em]
	\end{bmatrix}\label{eq:P}
\end{align}
where $\textbf{P} \in \mathbb{R}^{(n_s+n_{\bm{\Omega}})\times (n_s+n_{\bm{\Omega}})}$ consists of symmetric submatrices $\textbf{P}_1 \in \mathbb{R}^{n_s\times n_s}$ and $\textbf{P}_4\in \mathbb{R}^{n_{\bm{\Omega}}\times n_{\bm{\Omega}}}$, with $\textbf{P}_2 \in \mathbb{R}^{n_s\times n_{\bm{\Omega}}} = \textbf{P}_3^\intercal$. Substituting \eqref{eq:P} and \eqref{eq:F} into \eqref{eq:CovEst} yields
\begin{align}
	\scalebox{.6}{$\lefteqn{\begin{bmatrix}
				\textbf{P}_1(i|i{-}1) & \textbf{P}_2(i|i{-}1) \\[0.3em]
				\textbf{P}_3(i|i{-}1) & \textbf{P}_4(i|i{-}1) \\[0.3em]
			\end{bmatrix} =}$} \nonumber\\
	&\scalebox{.6}{$\begin{bmatrix}
			\textbf{F}_1(i) & \textbf{F}_2(i) \\[0.3em]
			\textbf{0} & \textbf{I}_{n_{\bm{\Omega}}} \\[0.3em]
		\end{bmatrix}	\begin{bmatrix}
			\textbf{P}_1(i{-}1|i{-}1) & \textbf{P}_2(i{-}1|i{-}1) \\[0.3em]
			\textbf{P}_3(i{-}1|i{-}1) & \textbf{P}_4(i{-}1|i{-}1) \\[0.3em]
		\end{bmatrix}\begin{bmatrix}
			\textbf{F}^\intercal_1(i) & \textbf{0} \\[0.3em]
			\textbf{F}^\intercal_2(i) & \textbf{I}_{n_{\bm{\Omega}}} \\[0.3em]
		\end{bmatrix} + \begin{bmatrix}
			\sigma^2_{s}\textbf{I}_{n_s} & \textbf{0} \\[0.3em]
			\textbf{0} & \sigma^2_{\bm{\Omega}}\textbf{I}_{n_{\bm{\Omega}}} \\[0.3em]
		\end{bmatrix}$}.
\end{align}
Using the superscripts $^-$ and $^+$ as shorthands for the \textit{a priori} estimate $(i|i{-}1)$ and the \textit{a posteriori} estimates $(i{-}1|i{-}1)$ or $(i|i)$, where appropriate, yields the following update rules:
\begin{align}
	\textbf{P}^{-}_1 &= \scalebox{.85}{$\left[\textbf{F}_1\textbf{P}^+_1 + \textbf{F}_2(\textbf{P}^+_2)^\intercal\right]\textbf{F}^\intercal_1 + \underbrace{\left[\textbf{F}_1\textbf{P}^{+}_2+\textbf{F}_2\textbf{P}^{+}_4\right]}_{\textbf{P}_2^-} \textbf{F}_2^\intercal+\sigma^2_s \textbf{I}_{n_s}\label{Eq:P1Min}$}\\
	\textbf{P}^{-}_2 &= \textbf{F}_1\textbf{P}^{+}_2+\textbf{F}_2\textbf{P}^{+}_4\label{eq:P2}\\
	\textbf{P}^{-}_3 &= \textbf{P}^{-\intercal}_2 \\
	\textbf{P}^{-}_4 &= \textbf{P}^{+}_4+\sigma^2_{\bm{\Omega}} \textbf{I}_{n_{\bm{\Omega}}}\label{Eq:P4Min}.
\end{align}

From the innovation covariance matrix
\begin{align}
	\mathds{S}_i = \textbf{H} \textbf{P}^-\textbf{H}^\intercal + \textbf{R} = \textbf{H} \textbf{P}^-\textbf{H}^\intercal + \sigma_y^2\textbf{I}_{n_y}
\end{align}
where $\textbf{R}$ is the measurement covariance matrix and $\sigma_y$ the output standard deviation, we obtain the Kalman gain $\mathds{K}$ as
\begin{align}
	\mathds{K}_i &= \textbf{P}^-\textbf{H}_i^\intercal \mathds{S}_i^{-1}\nonumber\\
	&=	\scalebox{.9}{$\begin{bmatrix}
			\textbf{P}_1^- & \textbf{P}_2^- \\[0.3em]
			\textbf{P}_3^- & \textbf{P}_4^- \\[0.3em]
		\end{bmatrix} \begin{bmatrix}
			\mathds{I}^\intercal \\[0.3em]
			\textbf{0} \\[0.3em]
		\end{bmatrix}\left(\begin{bmatrix}
			\mathds{I} & \textbf{0}\\[0.3em]
		\end{bmatrix}\begin{bmatrix}
			\textbf{P}_1^- & \textbf{P}_2^- \\[0.3em]
			\textbf{P}_3^- & \textbf{P}_4^- \\[0.3em]
		\end{bmatrix} \begin{bmatrix}
			\mathds{I}^\intercal \\[0.3em]
			\textbf{0} \\[0.3em]
		\end{bmatrix} + \sigma_y^2\textbf{I}_{n_y} \right)^{-1}$}\nonumber\\
	&= \begin{bmatrix}
		\textbf{P}_1^-\mathds{I}^\intercal \\[0.3em]
		\textbf{P}_3^-\mathds{I}^\intercal \\[0.3em]
	\end{bmatrix}\left(\mathds{I}\textbf{P}_1^-
	\mathds{I}^\intercal + \sigma_y^2\textbf{I}_{n_y} \right)^{-1}\nonumber\\
	&= \begin{bmatrix}
		\textbf{P}_1^-\mathds{I}^\intercal \\[0.3em]
		\left(\textbf{P}_2^-\right)^\intercal\mathds{I}^\intercal \\[0.3em]
	\end{bmatrix}\left(\mathds{I}\textbf{P}_1^-
	\mathds{I}^\intercal + \sigma_y^2\textbf{I}_{n_y} \right)^{-1}\nonumber\\
	&\stackrel{\Delta}{=} \begin{bmatrix}
		\textbf{L}_1 \\[0.3em]
		\textbf{L}_2 \\[0.3em]
	\end{bmatrix}\left(\textbf{M}_i + \sigma_y^2\textbf{I}_{n_y} \right)^{-1}\nonumber \stackrel{\Delta}{=} \begin{bmatrix}
		\textbf{L}_1 \\[0.3em]
		\textbf{L}_2 \\[0.3em]
	\end{bmatrix}\textbf{N}_i
\end{align}
where the following matrices store intermediate results:
\begin{align}
	\textbf{L}_1 &\stackrel{\Delta}{=} \textbf{P}^-_1\mathds{I}^\intercal\label{Eq:L1}\\
	\textbf{L}_2 &\stackrel{\Delta}{=} (\textbf{P}^-_2)^\intercal\mathds{I}^\intercal\label{Eq:L2}\\
	\textbf{M} &\stackrel{\Delta}{=} \mathds{I}\textbf{L}_1 \label{Eq:M}\\
	\textbf{N} &\stackrel{\Delta}{=} (\textbf{M}+\sigma^2_y \textbf{I}_{n_y})^{-1}\label{Eq:N}
\end{align}
with $\textbf{L}_1\in\mathbb{R}^{n_s\times n_y}$ (last $n_y$ columns of $\textbf{P}_1^-$), $\textbf{L}_2\in\mathbb{R}^{n_{\bm{\Omega}}\times n_y}$ (last $n_y$ rows of $\textbf{P}_2^-$, transposed), and $\textbf{M}_i\in\mathbb{R}^{n_y\times n_y}$ (the $n_y\times n_y$ lower-right corner of $\textbf{P}_1^-$) being submatrices of the decomposed state-covariance matrix $\textbf{P}$, since the linear mappings $\textbf{H}$ and $\mathds{I}$ are defined in \eqref{eq:H} as simple projections onto the last $n_y$ coordinates of state $\textbf{s}_i$, and $\textbf{N}$ is the inverse of the innovation covariance matrix.

Clearly, the Kalman gain matrix consists of two parts
\begin{align}
	\mathds{K} = \begin{bmatrix}
		\mathds{K}_1 \\[0.3em]
		\mathds{K}_2 \\[0.3em]
	\end{bmatrix}=\begin{bmatrix}
		\textbf{L}_1\textbf{N} \\[0.3em]
		\textbf{L}_2\textbf{N}\\[0.3em]
	\end{bmatrix}\label{Eq:K}
\end{align}
where $\mathds{K}_1\in\mathbb{R}^{n_s\times n_y}$ describes the changes in network activity $\textbf{s}_i$, and $\mathds{K}_2\in\mathbb{R}^{n_{\bm{\Omega}}\times n_y}$ corresponds to changes in the weights $\bm{\Omega}_i$, in response to errors $\textbf{e}$. 

Updating the \textit{a posteriori} state estimate gives
\begin{align}
	\mathbbm{s}^+ &= \mathbbm{s}^- + \mathds{K}\textbf{e}\nonumber\\
	\begin{bmatrix}
		\textbf{s} \\[0.3em]
		\bm{\Omega}\\[0.3em]
	\end{bmatrix}^+ &=\begin{bmatrix}
		\textbf{s} \\[0.3em]
		\bm{\Omega}\\[0.3em]
	\end{bmatrix}^- +
	\begin{bmatrix}
		\mathds{K}_1 \\[0.3em]
		\mathds{K}_2 \\[0.3em]
	\end{bmatrix}\textbf{e}.
\end{align}

Updating the \textit{a posteriori} covariance estimate gives
\begin{align}
	\textbf{P}^+ &= (\textbf{I} - \mathds{K}\textbf{H})\textbf{P}^-\nonumber\\
	\begin{bmatrix}
		\textbf{P}_1^+ & \textbf{P}_2^+ \\[0.3em]
		\textbf{P}_3^+ & \textbf{P}_4^+ \\[0.3em]
	\end{bmatrix} &= \left(\textbf{I} - \begin{bmatrix}
		\mathds{K}_1 \\[0.3em]
		\mathds{K}_2 \\[0.3em]
	\end{bmatrix}\begin{bmatrix}
		\mathds{I} & \textbf{0}\\[0.3em]
	\end{bmatrix}\right)\begin{bmatrix}
		\textbf{P}_1^- & \textbf{P}_2^- \\[0.3em]
		\textbf{P}_3^- & \textbf{P}_4^- \\[0.3em]
	\end{bmatrix} \nonumber\nonumber\\
	&=\begin{bmatrix}
		\textbf{P}_1^- & \textbf{P}_2^- \\[0.3em]
		\textbf{P}_3^- & \textbf{P}_4^- \\[0.3em]
	\end{bmatrix} - \begin{bmatrix}
		\mathds{K}_1 \\[0.3em]
		\mathds{K}_2 \\[0.3em]
	\end{bmatrix}\begin{bmatrix}
		\mathds{I}\textbf{P}_1^- & \mathds{I}\textbf{P}_2^-\\[0.3em]
	\end{bmatrix}\nonumber\\
	&=\begin{bmatrix}
		\textbf{P}_1^- & \textbf{P}_2^- \\[0.3em]
		\textbf{P}_3^- & \textbf{P}_4^- \\[0.3em]
	\end{bmatrix} - \begin{bmatrix}
		\mathds{K}_1\textbf{L}^\intercal_1 & \mathds{K}_1\textbf{L}^\intercal_2\\[0.3em]
		\mathds{K}_2\textbf{L}^\intercal_1 & \mathds{K}_2\textbf{L}^\intercal_2\\[0.3em]
	\end{bmatrix}.
\end{align}

Specifically, measurement updates are given by
\begin{align}
	\textbf{s}^+ &= \textbf{s}^- + \mathds{K}_1 \textbf{e}\label{Eq:SPlus}\\
	\bm{\Omega}^+ &= \bm{\Omega}^- + \mathds{K}_2 \textbf{e}\label{Eq:OmegaPlus}\\
	\textbf{P}^{+}_1 &= \textbf{P}^{-}_1 - \mathds{K}_1 \textbf{L}_1^\intercal\label{Eq:P1Plus}\\
	\textbf{P}^{+}_2 &= \textbf{P}^{-}_2 - \mathds{K}_1 \textbf{L}_2^\intercal\label{Eq:P2Plus}\\
	\textbf{P}^{+}_4 &= \textbf{P}^{-}_4 - \mathds{K}_2 \textbf{L}_2^\intercal.\label{Eq:P4Plus}
\end{align}
The covariance blocks are initialized as follows
\begin{align}
	\textbf{P}_1(0) &= \sigma^2_s \textbf{I}_{n_s}\\
	\textbf{P}_4(0) &= \sigma^2_{\bm{\Omega}} \textbf{I}_{n_{\bm{\Omega}}}\label{Eq:P4Init}\\
	\textbf{P}_2(0) &= \textbf{0}\label{eq:P2Initial}
\end{align}
where diagonal matrices are used for simplicity, and the initial state is assumed to be independent of the filter weights. The explicit Hilbert space functional Bayesian filter (expFBF) algorithm is summarized in Algorithm \ref{alg:expFBF}. Detailed derivation of the Jacobians $\textbf{F}_1(i)$ and $\textbf{F}_2(i)$ is presented in the Appendix.

\begin{algorithm}
	\caption{Explicit Hilbert Space Functional Bayesian Filter}\label{alg:expFBF}
	\begin{algorithmic}[h]
		\State \textbf{Initialization:}
		\State $n_u$: input dimension
		\State $n_y$: output dimension
		\State $n_s$: state dimension
		\State $a_u$: input kernel parameter
		\State $a_s$: state kernel parameter
		\State $\widetilde{\phi}(\cdot)$ input feature map
		\State $\widetilde{\psi}(\cdot)$ state feature map
		\State $n_{\bm{\Omega}}$: feature-space dimension
		\State $\sigma^2_s$: state variance
		\State $\sigma^2_y$: output variance
		\State $\sigma^2_{\bm{\Omega}}$: weight variance
		\State $\textbf{P}_1(0) = \sigma^2_s \textbf{I}_{n_s}$
		\State $\textbf{P}_4(0) = \sigma^2_{\bm{\Omega}}\textbf{I}_{n_{\bm{\Omega}}}$
		\State $\textbf{P}_2(0) = \textbf{0}$
		\State Randomly initialize state $\textbf{s}_{0}$
		\State Randomly initialize weights $\bm{\Omega}$: matrices $\textbf{A}$ and $\textbf{B}$
		\State $\mathds{I}=\begin{bmatrix}
				\textbf{0} & \textbf{I}_{n_y} \\[0.3em]
			\end{bmatrix}\in\mathbb{R}^{n_y\times n_s}$: measurement matrix\\
		\For{$i = 1,\cdots$}
		\State 		\textbf{\underline{Predict}:}
		\State 		Get current input $\textbf{u}_i$ and past state $\textbf{s}_{i-1}$
		\State	Map input $\textbf{u}_i$ to $\widetilde{\phi}(\textbf{u}_i)$
		\State	Map state $\textbf{s}_i$ to $\widetilde{\psi}(\textbf{s}_i)$
		\State 		Propagate \textit{a priori} state estimate
		\State 	\quad$\textbf{s}^-_{i} = \textbf{A}_{i}\widetilde{\psi}(\textbf{s}_{i-1}) + \textbf{B}_{i} \widetilde{\phi}(\textbf{u}_{i})$\hfill \eqref{eq:exp_state_transition}
		\State 		Compute state transition dynamics:
		\State \quad$\textbf{F}_1(i) = \frac{\partial \textbf{s}_i}{\partial \textbf{s}_{i-1}}$\hfill \eqref{Eq:F1}
		\State 		\quad$\textbf{F}^{(\textbf{A})}_2(i) =\frac{\partial \textbf{s}_i}{\partial \textbf{A}(i)}$\hfill \eqref{Eq:F2A}
		\State 		\quad$\textbf{F}^{(\textbf{B})}_2(i) = \frac{\partial \textbf{s}_i}{\partial \textbf{B}(i)}$\hfill \eqref{Eq:F2B}
		\State 		Propagate \textit{a priori} covariance estimate:\\
		\State 		\quad$\textbf{P}^{-}_1 = \left[\textbf{F}_1\textbf{P}^+_1 + \textbf{F}_2(\textbf{P}^+_2)^\intercal\right]\textbf{F}^\intercal_1$
		\State \quad$\quad \qquad + \underbrace{\left[\textbf{F}_1\textbf{P}^{+}_2+\textbf{F}_2\textbf{P}^{+}_4\right]}_{\textstyle \textbf{P}_2^-} \textbf{F}_2^\intercal + \sigma^2_s\textbf{I}_{n_s}$\hfill \eqref{Eq:P1Min}
		\State 	\quad$\textbf{P}^{-}_4 = \textbf{P}^{+}_4+\sigma^2_{\bm{\Omega}} \textbf{I}_{n_{\bm{\Omega}}}$\hfill \eqref{Eq:P4Min}
		\State 	\textbf{\underline{Update}:}
		\State 		Obtain innovation or measurement residual
		\State 		\quad$\textbf{e}_i = \textbf{d}_i - \textbf{y}_i $
		\State 		Compute Kalman gain:
		\State \quad$\textbf{L}_1 \leftarrow$ last $n_y$ columns of $\textbf{P}_1^-$\hfill \eqref{Eq:L1}
		\State 	\quad	$\textbf{L}_2 \leftarrow$ last $n_y$ columns of $(\textbf{P}_2^-)^\intercal$\hfill \eqref{Eq:L2}
		\State 		\quad$\textbf{M}\leftarrow$ the $n_y\times n_y$ lower-right corner of $\textbf{P}_1^-$\hfill \eqref{Eq:M}
		\State   \quad	$\textbf{N}\leftarrow (\textbf{M}+\sigma^2_y \textbf{I}_{n_y})^{-1}$\hfill \eqref{Eq:N}
		\State 			\quad$\mathds{K}_1\leftarrow \textbf{L}_1 \textbf{N}$\hfill \eqref{Eq:K}
		\State 		\quad$\mathds{K}_2\leftarrow \textbf{L}_2 \textbf{N}$\hfill \eqref{Eq:K}
		\State 		Update \textit{a posteriori} estimates:
		\State 		\quad$\textbf{s}^+ = \textbf{s}^- + \mathds{K}_1 \textbf{e}$\hfill \eqref{Eq:SPlus}
		\State 		\quad$\bm{\Omega}^+ = \bm{\Omega}^- + \mathds{K}_2 \textbf{e}$\hfill \eqref{Eq:OmegaPlus}
		\State 		\quad$\textbf{P}^{+}_1 = \textbf{P}^{-}_1 - \mathds{K}_1 \textbf{L}_1^\intercal$\hfill \eqref{Eq:P1Plus}
		\State 		\quad$\textbf{P}^{+}_2 = \textbf{P}^{-}_2 - \mathds{K}_1 \textbf{L}_2^\intercal$\hfill \eqref{Eq:P2Plus}
		\State 			\quad$\textbf{P}^{+}_4 = \textbf{P}^{-}_4 - \mathds{K}_2 \textbf{L}_2^\intercal$\hfill \eqref{Eq:P4Plus}
		\EndFor
	\end{algorithmic}
\end{algorithm}

\subsection{Relationship to the Koopman Operator} \label{Sec:Koopman}
Koopman originally formulated the linear operator as a discrete-time mapping for Hamiltonian systems \cite{Koopman31}.  The Koopman operator can be generalized as follows:

\begin{defn}[Koopman Operator]
For a continuous-time dynamical system\index{dynamical system}
\begin{align}
	\frac{d{\bf x}}{dt} = \bf{f}({\bf x}) \label{eq:Ukoop}
\end{align}
where ${\bf x}\in\mathcal{X}$ is the state on a smooth $n_x$-dimensional manifold $\mathcal{X}$.  The Koopman operator ${\cal K}$ is an infinite-dimensional linear operator that acts on all observable functions $g:\mathcal{X}\rightarrow\mathbb{C}$ such that
\begin{align}
{\cal K} {g}({\bf x})  = { g}\left( {\bf f}({\bf x}) \right) = g\circ {\bf f}({\bf x}).\label{eq:koop}
\end{align}
\end{defn}
By design, the Koopman operator is a linear, infinite-dimensional operator that acts on the Hilbert space $\mathcal{H}$ of all scalar measurement functions $g$. Similarly, the weight $\bm{\Omega}$ we learn in the RKHS acts on functions of the state space of the dynamical system,  trading \textit{nonlinear finite}-dimensional dynamics for \textit{linear infinite}-dimensional dynamics. The two operators are theoretically equivalent. The Koopman perspective offers two compelling advantages: nonlinear problems can be addressed utilizing standard linear operator theory and spectral decomposition. In practice, the computation of the Koopman operator requires a finite-dimensional representation, by considering a sufficiently large yet finite sum of modes to approximate the Koopman spectra. Crucial to numerical implementation of this definition is understanding how to select a finite set of observables $g({\bf x})$, which remains an ongoing challenge. 

The ML kernel perspective offers two significant benefits:
\begin{enumerate}
	\item The Hilbert space can be constructed deterministically, agnostic to the nonlinear dynamics. The mappings induce a positive
	definite kernel function satisfying Mercer’s conditions under the closure properties (where, positive-definite kernels are closed under addition,
	multiplication, and scaling). The Gaussian kernel has universal approximation property: it approximates uniformly an arbitrary continuous target function over any compact domain.
	\item Bayesian filter can adaptively learn the Koopman operator that is minimum variance, allowing the system to continuously track the changes across an extended period of time, ideally suited for modern data-driven applications such as real-time ML using streaming data. 
\end{enumerate}
For practical applications, we can define a finite-dimensional FBF using polynomial approximation of the universal kernel.

\section{Experiments and Results}\label{Sec:Results}
Here, we illustrate and evaluate the proposed explicit Hilbert space Functional Bayesian Filter using numerical examples. As a proof-of-concept, we consider the following tasks: chaotic time series estimation and modeling nonlinear PDE.

\subsection{Cooperative Filtering for Signal Enhancement}
First, we consider the scenario of an unknown nonlinear dynamical system with only noisy observations available, and compare the denoising performances (signal enhancement) between the functional Bayesian filter \cite{FBF2022}, its finite-dimensional counterpart the expFBF, and the cubature Kalman filter (CKF) \cite{Arasaratnam2009} on the Mackey-Glass (MG) chaotic time series $y_t$ \cite{Mackey77}, defined by the following delay differential equation
\begin{align*}
	\frac{d y_t}{d t} =\frac{\beta y_{(t-\tau)}}{1+y^{n}_{(t-\tau)}} -\gamma y_t
\end{align*}
where $\beta=0.2$, $\gamma=0.1$, $\tau=30$, $n=10$, discretized at a sampling period of 6 seconds using the fourth-order Runge-Kutta method, with initial condition $y_0 = 0.9$. Chaotic dynamics are highly sensitive to initial conditions, where even a small change in the current state can lead to vastly different outcomes over time. This makes long-term prediction intractable and is commonly known as the butterfly effect \cite{Ott02}. 

Cooperative filtering seeks to build an empirical model by extracting (pseudo-) clean data from the noisy measurements. For unknown dynamics, the signal estimator is coupled with the weight parameter estimator. The cubature Kalman approach (specifically, the square-root version or SCKF) uses a recurrent neural network (RNN) to model the dynamics: the weights of the RNN are estimated from the latest signal estimate and \textit{vice versa}.   The SSM for the RNN architecture, trained using SCKF, is constructed as
\begin{align*}
	\textbf{W}_i &= \textbf{W}_{i-1} + \bm{q}_{i-1}\\
	d_i &= \textbf{W}^{(\rm o)} {\rm h}(\textbf{W}^{(\rm r)}\textbf{x}_{i-1}+\textbf{W}^{(\rm i)}\textbf{u}_i) +r_i 
\end{align*}
where the input weight $\textbf{W}^{(\rm i)}$, the recurrent weight $\textbf{W}^{(\rm r)}$, and the output weight $\textbf{W}^{(\rm o)}$ are matrices of appropriate dimensions (arranged into an orderly weight vector $\textbf{W}_i$). The process noise is zero-mean Gaussian with covariance $Q_{i}$, i.e., $\bm{q}_i\sim \mathcal{N}(0,Q_{i})$, and the measurement noise is $r_i\sim \mathcal{N}(0,R_{i})$. The input $\textbf{u}_i = [u_i, u_{i-1}, \cdots, u_{i-(\ell-1)}]$ has embedding dimension $\ell = 7$, and the self-recurrent hidden layer contains 5 neurons ($\textbf{x}_i \in \mathbb{R}^5$). The hidden layer activation uses the hyperbolic tangent function, while the single output neuron is linear ($d_i\in \mathbb{R}$). The 7-5R-1 RNN consists of 71 weight parameters including bias terms.

\begin{figure}[t!]
	\centering
	\includegraphics[width=0.3\textwidth]{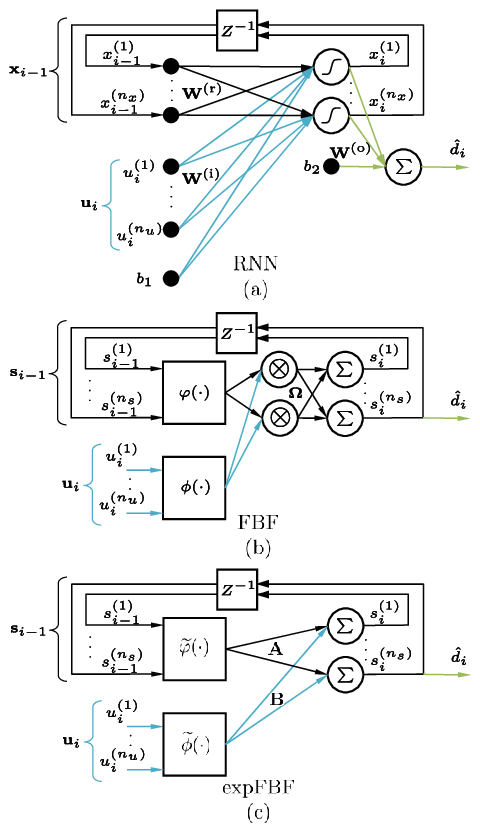}
	\caption{Recurrent network trained using (a) Square-Root Cubature Kalman Filter and RNN (b) Functional Bayesian Filter using the tensor-product kernel (c) Explicit Hilbert space FBF using a sum kernel.}
	\label{fig:model_compare}
\end{figure}

For the FBF and the expFBF, we construct linear models with the same state and input dimensions as the 7-5R-1 RNN, but with more parsimonious architectures. Fig. \ref{fig:model_compare} compares the constructions for the three approaches. The kernel parameters for the state, input, state covariance $\textbf{P}_1$, and weight covariance $\textbf{P}_4$ are $a_s = 0.6$, $a_u=1.8$, $a_{P_1}= 0.4$, and $a_{P_4} = 0.2$, respectively. The state covariance, output variance, and weight covariance are initialized as $\sigma^2_{s} = \sigma^2_{y} = 0.09$, $\sigma^2_{P_4} = 10$, respectively. There are no bias terms for our FBF and expFBF implementations. Using the small-step-size theory framework \cite{Haykin02}, which self-regularizes KAF, we scale the state Kalman gain $\mathds{K}_1$ by a constant factor of 0.4, and the weight gain $\mathds{K}_2$ by a constant factor of 0.1. Note, the FBF is an online kernel method with a dictionary constructed incrementally with each incoming sample (the initial point is randomly initialized), e.g., after 100 samples, the dictionary size becomes 101. The expFBF is of a fixed construction. We set the number of term for the Taylor series expansion to 4, resulting in $n_{\widetilde{\psi}(s)} = \binom{5+4}{4} = 126$, and $n_{\widetilde{\phi}(u)} = \binom{7+4}{4} = 330$. 

A noisy (10 dB SNR) MG chaotic time sequence of 1000 samples is used for training. For each training run, ten batches were made. Each batch consists of 100 time-step updates, from a randomly selected starting point in the training sequence. The state of the RNN at time step $i=0$ was assumed to be zero, i.e., $\textbf{x}_0=\textbf{0}$. During the test phase, an independent sequence of 100 noisy samples is used with the network (SSM) weights fixed.

\begin{figure}[t!]
	\centering
	\begin{subfigure}
		\centering
		\includegraphics[width=0.4\textwidth]{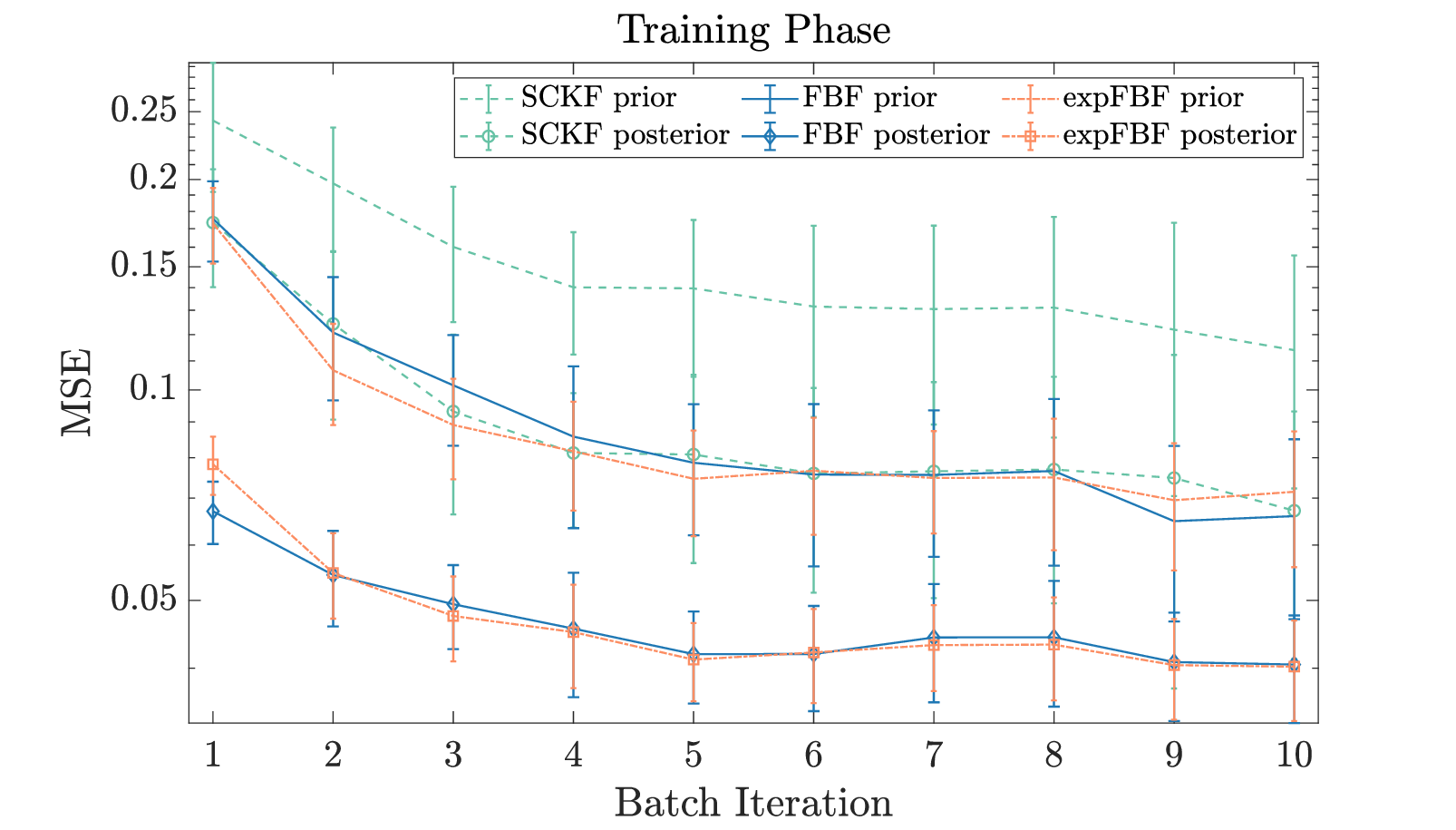}
	\end{subfigure}%
	\begin{subfigure}
		\centering
		\includegraphics[width=0.4\textwidth]{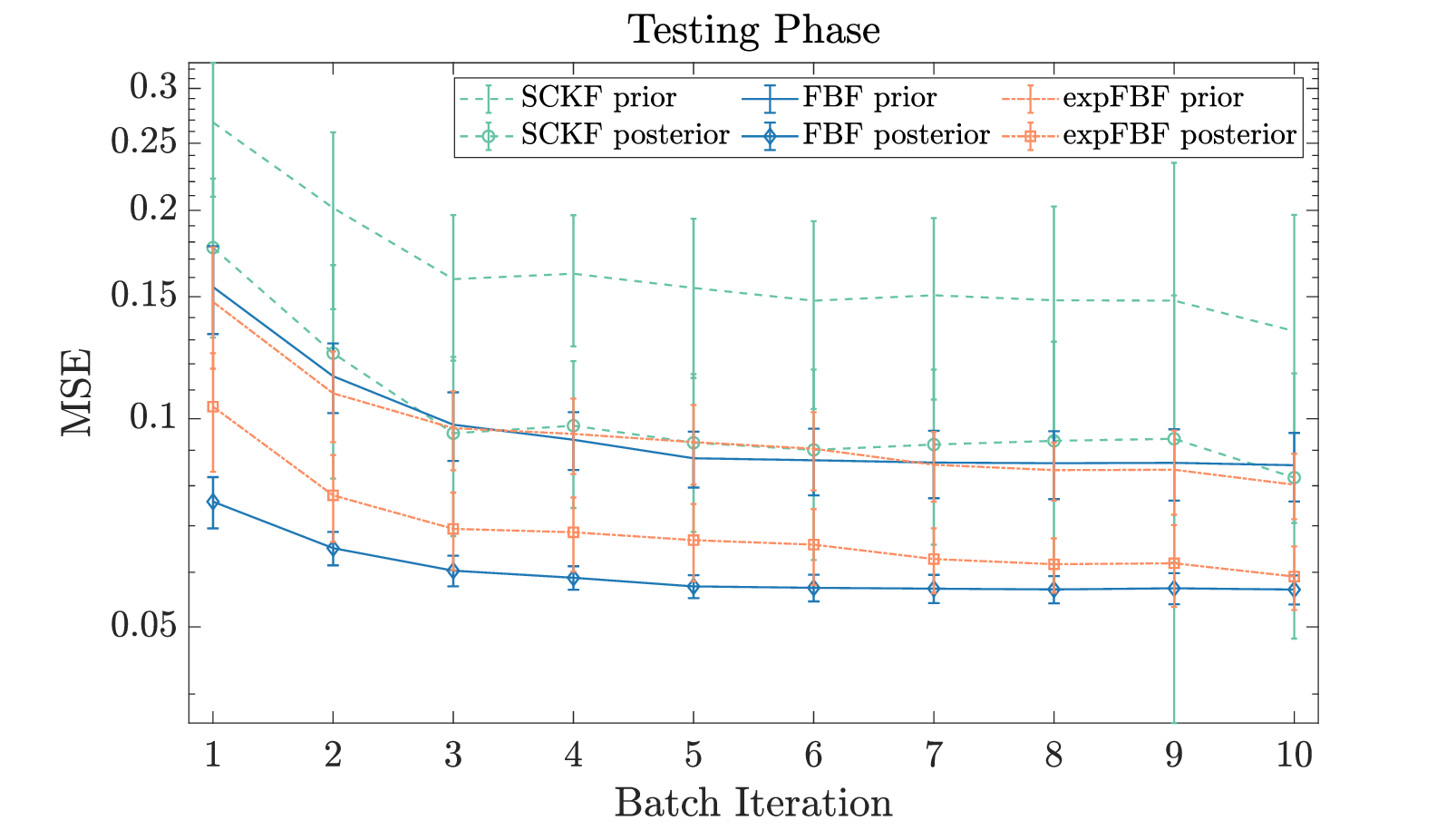}
	\end{subfigure}
	\caption{Ensemble-averaged Mean-Squared Error (MSE) over 50 runs vs. number of batch iterations (each training iteration consists of a 100-sample sequence with random starting point).}
	\label{fig:MGperformance}
\end{figure}

SCKF has been successfully validated to significantly outperform other known nonlinear filters such as EKF and central-difference Kalman filter (CDKF) and provides improved numerical stability over CKF \cite{Arasaratnam2009}. It is important to reiterate that the two essential properties of error covariance matrix (symmetry and positive definiteness) are always preserved in FBF, since we are using positive definite kernel functions satisfying Mercer's conditions, unlike input-space arithmetic operations such as CKF, where these two properties are often lost or destroyed and a square-root version is preferred. 

Fig. \ref{fig:MGperformance} shows the ensemble-averaged mean square error (MSE) over 50 independent runs versus the number of batch iterations (error bars represent one standard deviation), where each training iteration consists of a 100-sample noisy sequence with random starting point in the 1000-sample training data. The ``prior'' label denotes the time update using the predictive density before receiving a new measurement; ``posterior'', the measurement update from the posterior density. Clearly, the FBF significantly improves the quality of the signal as compared to the SCKF, and the expFBF approximates the performance of the FBF but with the added advantage of a finite-dimension construction. The FBF memory and computational complexities for each recursive
update are $O(n)$ and $O(n^2)$, respectively, where $n$ is the length of the data sequence. This can be quite prohibitive for continuous tracking, and would need to be paired with an appropriate sparsification technique to limit the number of data points or size of the basis used.

\subsection{Nonlinear Schr\"odinger Equation}

Next, we consider the expFBF and Koopman operator applied to a canonical second-order nonlinear PDE:  the nonlinear Schr\"odinger (NLS) equation
\begin{align}
	i  u_t + \frac{1}{2} u_{xx} + |u|^2 u =0
	\label{eq:nls_koop}
\end{align}
where $u(x,t)$ is a function of space and time, with many applications in theoretical physics, such as modeling the propagation of light in nonlinear optical fields or small-amplitude gravity waves on the surface of deep inviscid fluid. This study expands on the analysis in \cite{Kutz2018}. By discretizing the spatial variable $x$, we can Fourier transform the solution in space and employ a fourth-order Runge-Kutta method to step the solution forward in time.

We can compute the DMD by collecting snapshots of the dynamics over a given time window.  Specifically, we consider simulations of the equation with initial data
\begin{align}
	u(x,0)= 2\mbox{sech} (x)
	\label{Eq:NLS_init}
\end{align}
over the time interval $t\in[0,\pi]$.  The numerical simulation consists of 21 snapshots of the real-part of the dynamics, yielding the input-output snapshot matrices ${\textbf X}= [\textbf{x}_0,\textbf{x}_1,\cdots,\textbf{x}_{m-1}]$ and ${\textbf X}'= [\textbf{x}_1,\textbf{x}_2,\cdots,\textbf{x}_m]$, with $m = 21$. Spatially, 32 points are uniformly sampled from $[-15,15]$, i.e., $\textbf{x}_i\in\mathbb{R}^{32}$. DMD is a batch method, while expFBF is a recursive estimator, iteratively predicates and updates the state and its uncertainty, one step at a time.

Four different sets of features or observables are considered:
\begin{enumerate}
	\item The observables used in the DMD reduction are simply the original state variables ${\bf x}=u(x,t)$ at discrete space and time steps, i.e., ${\bf g}_{\mbox{\tiny DMD}} ({\bf x})= {\bf x}$.
	\item Koopman theory allows a broader set of observables. Here we consider the following sets:
	\begin{align*}
		{\bf g}_{k1}({\bf x}) &=\left[  \begin{array}{cc} {\bf x}\\ |{\bf x}|^2 {\bf x} \end{array} \right], \text{ and }
		{\bf g}_{k2}({\bf x}) &=\left[  \begin{array}{cc} {\bf x}\\ |{\bf x}|^2 \end{array}  \right].
	\end{align*}
	Both require different levels of knowledge or understanding of the underlying latent equations in the dynamical system: the first set of observables ${\bf g}_1({\bf x})$ uses the exact form of the nonlinearity in the NLS equation, which we denote by \textit{good} embedding. The second, ${\bf g}_2({\bf x})$, is chosen be a more generic quadratic nonlinearity, but uses the absolute value, which is important for the NLS equation due to the nonlinearity in the phase evolution. For the NSL equation, the observables in ${\bf g}_2({\bf x})$ are considered inferior to either the DMD or the judiciously selected ${\bf g}_1({\bf x})$ for the Koopman reconstruction since its nonlinearity does not match that of the dynamics under consideration, which we will refer to as \textit{bad} embedding \cite{Kutz2018}. Furthermore, we use a  reduced-order linear model constructed by a rank-$r$ truncation of the dominant eigenfunctions, with $r=10$. 
	\item The theory of RKHS enables universal kernel features. Here, we consider the Gaussian quadrature features ${\bf g}_{\rm GQ}({\bf x}) = \widetilde{\phi}({\bf x})$ or GQ embedding to approximate the Gaussian kernel (universal approximation property). These observables are agnostic to the state dynamics of the NLS and are instead associated with Mercer kernels.
\end{enumerate} 

\begin{figure}[t]
	\centering
	\includegraphics[width=0.34\textwidth]{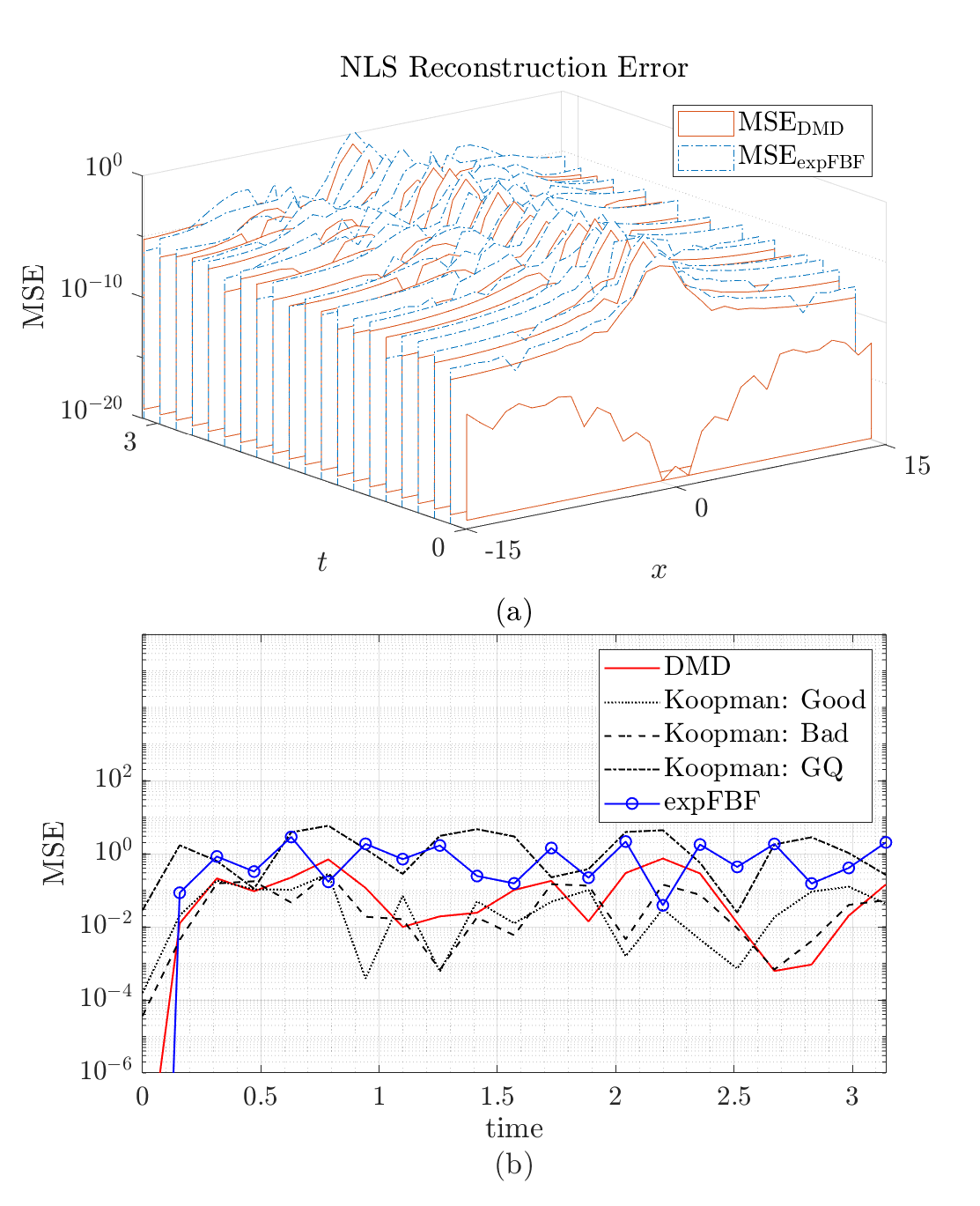}
	\caption{Short-term future state reconstruction errors of the NLS dynamics using a standard DMD approximation ${\bf g}_{\mbox{\tiny DMD}} ({\bf x})$, the NLS motivated observables ${\bf g}_1({\bf x})$, the generic quadratic observables ${\bf g}_2({\bf x})$, the Gaussian quadrature observable, and the explicit Hilbert space FBF using GQ observable: (a) the NLS reconstruction errors for DMD and expFBF, (b) the MSE summed across space for the four reconstructions.}
	\label{fig:short_term_NLS}
\end{figure}

\begin{figure}[t]
	\centering
	\includegraphics[width=0.34\textwidth]{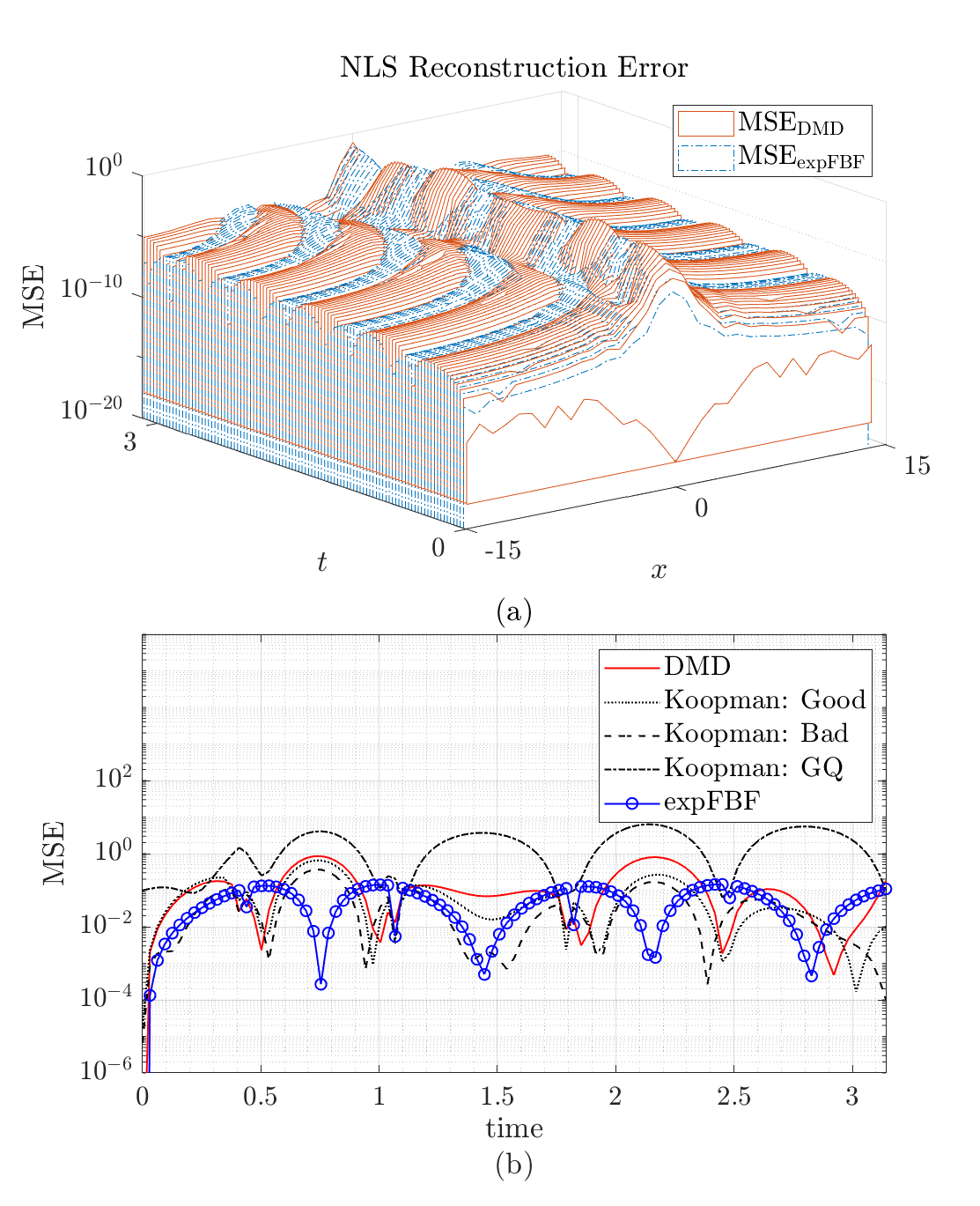}
	\caption{Long-term future state reconstruction errors of the NLS dynamics using a standard DMD approximation ${\bf g}_{\mbox{\tiny DMD}} ({\bf x})$, the NLS motivated observables ${\bf g}_1({\bf x})$, the generic quadratic observables ${\bf g}_2({\bf x})$, the Gaussian quadrature observable, and the explicit Hilbert space FBF using GQ observable: (a) the NLS reconstruction errors for DMD and expFBF, (b) the MSE summed across space for the four reconstructions.}
	\label{fig:long_term_NLS}
\end{figure}

The reconstruction errors of the NLS dynamics are shown in Fig. \ref{fig:short_term_NLS}. We can see that the choice of observables can significantly impact the performance of the Koopman approximation. Leveraging knowledge of the analytic solution, a judiciously selected observables can outperform generic observables and observables restricted to the original state variables. However, in practice, it is rarely the case that such linearizing transformations are known \textit{a priori}, let alone knowing the exact equations of the dynamics, to take advantage of the Koopman theory. The GQ observables performed the worst, since it is constructed not for the NLS dynamics, but rather induces an RKHS that approximates uniformly an arbitrary function over any compact domain. Even in this simple example, we see that a model-based method in the RKHS, such as the expFBF, can outperform the Koopman theoretic approximation using the same GQ observables.

\begin{figure}[t]
	\centering
	\includegraphics[width=0.37\textwidth]{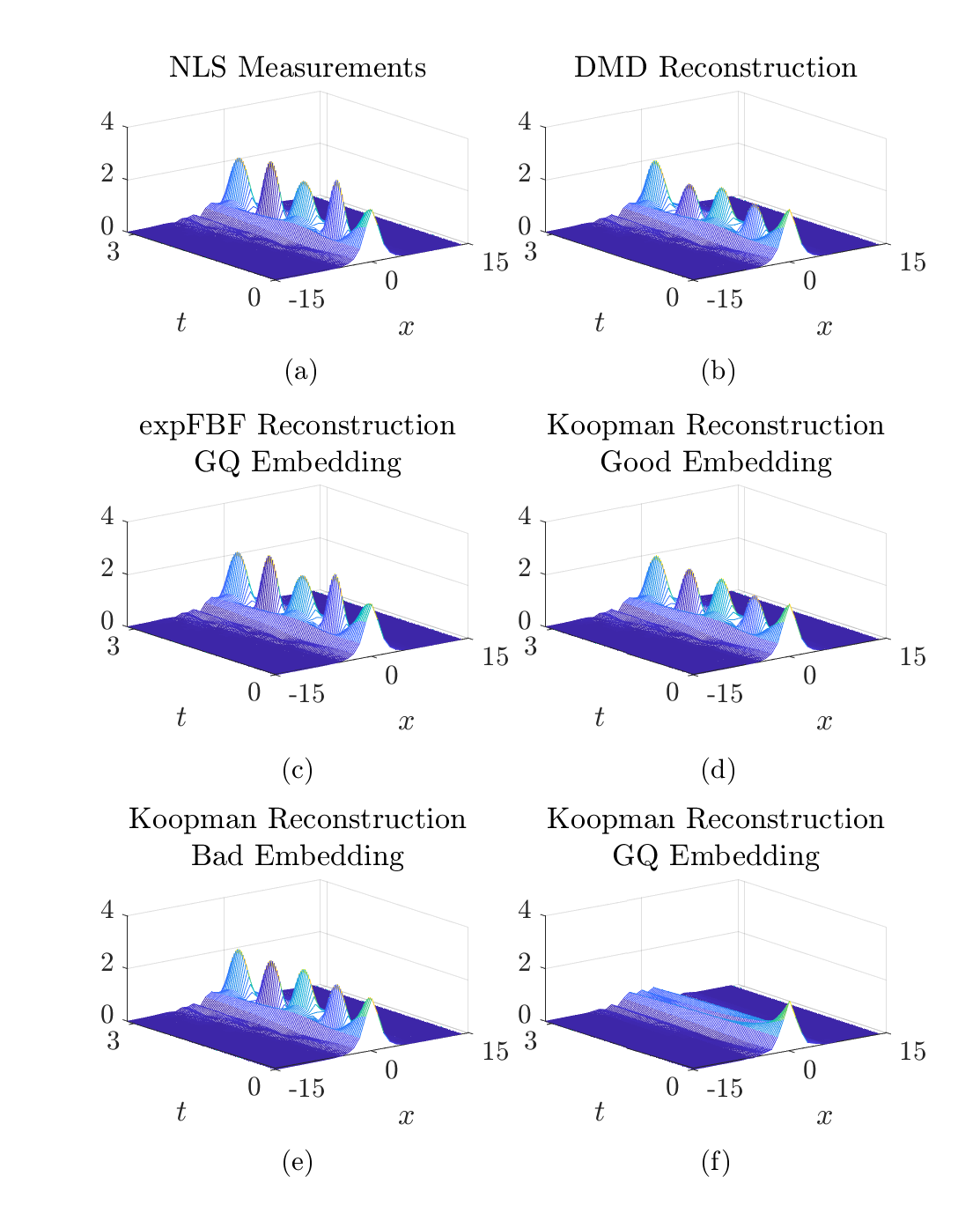}
	\caption{Long-term future state reconstructions of second-order nonlienar PDE using initial condition $u(x,0)= 2\mbox{sech} (x)$: (a) NLS measurements, (b) standard DMD approximation ${\bf g}_{\mbox{\tiny DMD}} ({\bf x})$, (c) expFBF, (d) the NLS motivated observables ${\bf g}_1({\bf x})$, (e) the generic quadratic observables ${\bf g}_2({\bf x})$, and (f) the Gaussian quadrature observable.}
	\label{fig:long_term_NLS_reconstruction}
\end{figure}

The expFBF is a learning algorithm that recursively updates the model parameters and the corresponding state estimations. To highlight this, we expand the number of snapshots in the NLS numerical simulation above from 21 to 101. The long-term prediction reconstruction errors of the NLS dynamics are shown in Fig. \ref{fig:long_term_NLS}. We see that given a longer period of adaptation, the expFBF outperforms the Koopman methods, even though the GQ observables are still the worst performing using the Koopman decomposition. The corresponding reconstructions are shown in Fig. \ref{fig:long_term_NLS_reconstruction}.

\begin{figure}[t]
	\centering
	\includegraphics[width=0.37\textwidth]{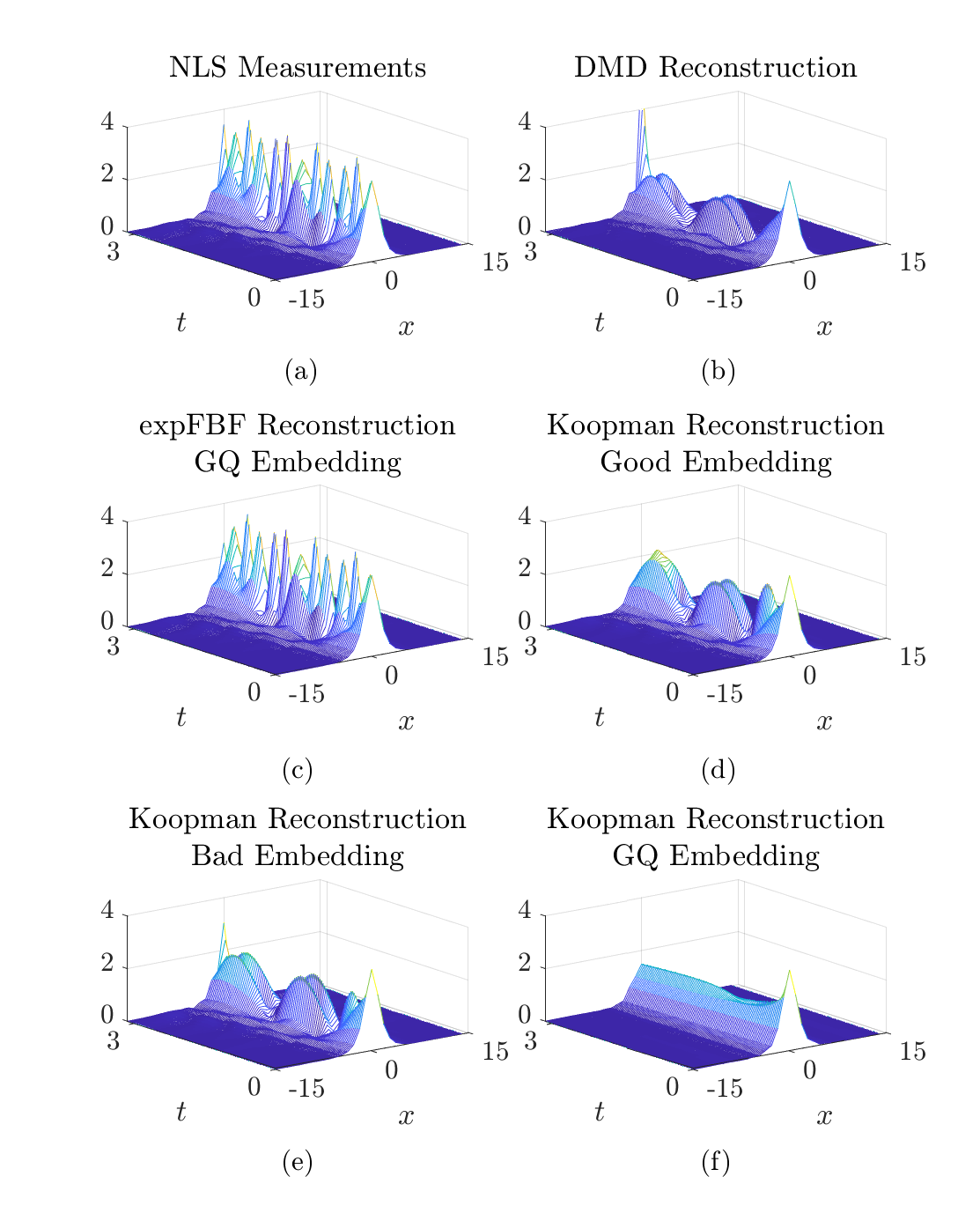}
	\caption{Long-term future state reconstructions of second-order nonlienar PDE using initial condition $u(x,0)= 3.1\mbox{sech} (x)$: (a) NLS measurements, (b) standard DMD approximation ${\bf g}_{\mbox{\tiny DMD}} ({\bf x})$, (c) expFBF, (d) the NLS motivated observables ${\bf g}_1({\bf x})$, (e) the generic quadratic observables ${\bf g}_2({\bf x})$, and (f) the Gaussian quadrature observable.}
	\label{fig:long_term_NLS_new_init}
\end{figure}

\begin{figure}[h]
	\centering
	\includegraphics[width=0.53\textwidth]{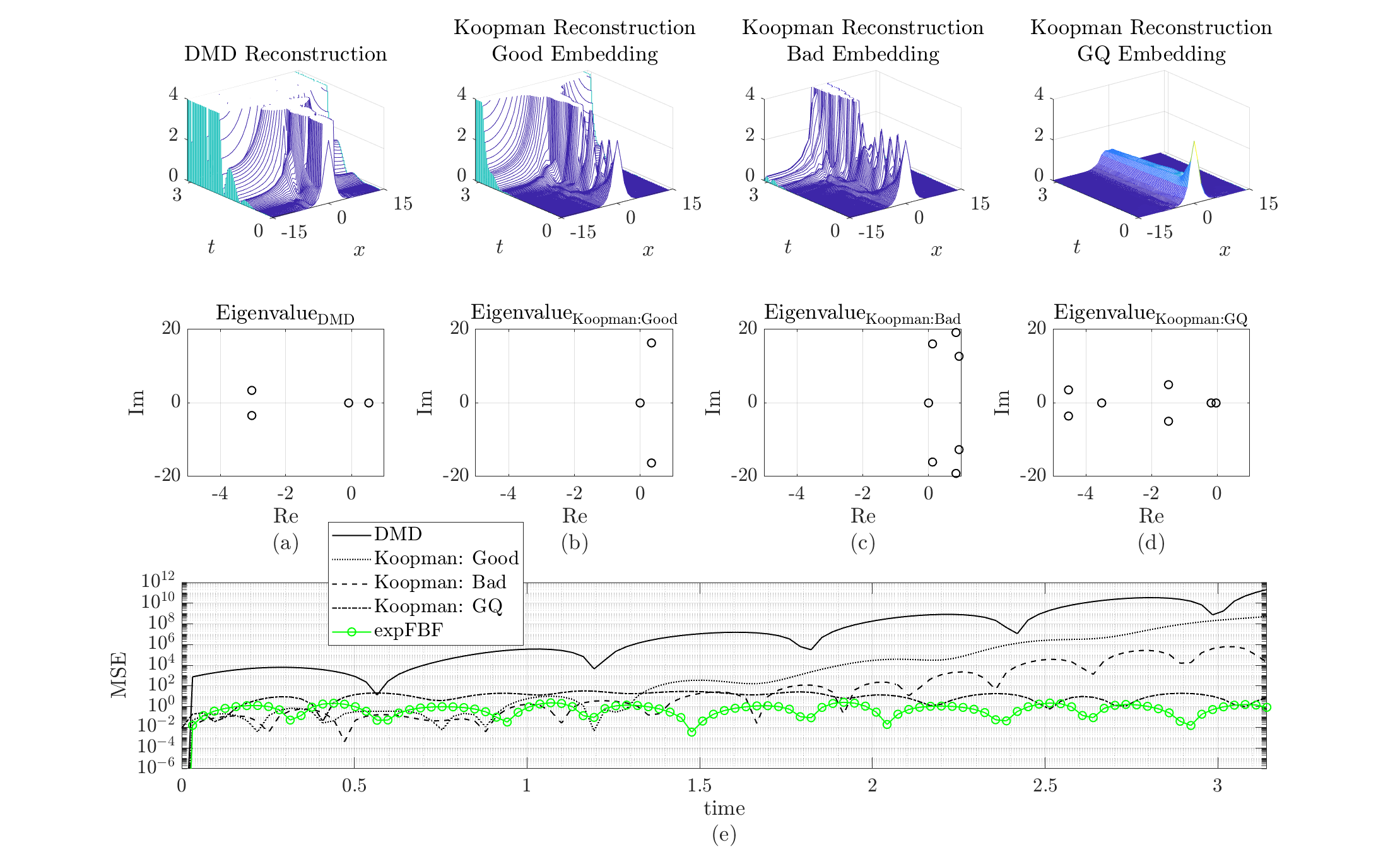}
	\caption{Reconstruction of the NLS dynamics using (a) a standard DMD approximation ${\bf g}_{\mbox{\tiny DMD}} ({\bf x})$,
		(b) the NLS motivated ${\bf g}_1({\bf x})$, (c) a quadratic observable ${\bf g}_2({\bf x})$ and (d) the GQ observables. The Koopman spectra
		for each observable is show in the second row. Note that the observable ${\bf g}_1({\bf x})$ produces a spectra which is approximately purely imaginary which is expected of the 2-soliton evolution. But here, we modified the initial condition by changing the scaling from 2 to 3.1. The NLS reconstruction MSE summed across space and plotted against time is shown in (e).
		\label{fig:long_term_NLS_new_init_r30}}
	\label{fig:RNN}
\end{figure}

Next, we investigate the performances' sensitivity to the initial condition. We modify the coefficient in \eqref{Eq:NLS_init} to
\begin{align}
	u(x,0)= 3.1\mbox{sech} (x).
	\label{eq:nlsi}
\end{align}
The reconstructions of the NLS dynamics for 101 time steps using the new initial condition is shown in Fig. \ref{fig:long_term_NLS_new_init}. We see that the dynamical structures are much more complex and nuanced here than before, and the low-rank DMD and Koopman reconstructions are not able to capture these spatial-temporal structures. Again, the Koopman reconstruction of the GQ embedding performed the worst, but by constructing a state-space model in the RKHS induced by the GQ observables, and using Bayesian updates, we are able to capture the spatiotemporal dynamics with high fidelity.

Fig. \ref{fig:long_term_NLS_new_init_r30} shows the performance when we increase the rank $r$ from 10 to 30. We see that the Koopman methods become unstable by considering more eigenfunctions, and the errors explode in time. The \textit{good} embedding becomes worse than the \textit{bad} embedding, since ${\bf g}_{k1}({\bf x})$ contains a cubic form vs. the quadratic nonlinearity in ${\bf g}_{k2}({\bf x})$. The Koopman reconstruction using GQ observables now has the best performance among the Koopman methods. This highlights the critical challenge in Koopman theory, selecting meaningful and robust observables. The extended DMD and kernel DMD method were also shown to be highly sensitive to observable selection in \cite{Kutz2018}.

Unlike the Koopman decomposition, whose effectiveness hinges almost exclusively on the selection of observables, the KAF operator-theoretic approach we propose here is robust and the feature space deterministically constructed, agnostic to the input dynamics. The Koopman eigenfunctions and eigenvalues capture the evolution of the linearized dynamics on a well-selected observable space while the GQ transformed states induces an RKHS that approximates the universal kernel. Observable selection for the NLS problem was facilitated by precise knowledge of the governing equation.  However, in many real-world applications, such expert knowledge is absent, and we must rely solely on data. This is precisely the appeal of the methodology we introduced here.

\section{Conclusion}\label{Sec:Conclusion}
The theory of RKHS is a powerful, versatile, and theoretically-grounded unifying framework to solve nonlinear problems in data-driven analysis such as signal processing and machine learning.	The standard approach relies on the \emph{kernel trick} to perform pairwise evaluations of a kernel function, which bypasses the explicit feature map (potentially infinite-dimensional) but leads to scalability issues for large datasets due to its linear and superlinear growth with respect to the size of the training data. In this paper, we proposed an explicit-space functional Bayesian filter to perform recursive state estimation using linear operators in a finite-dimensional RKHS that approximate the performance of a universal kernel for nonlinear dynamical systems, defined by polynomials such as the Gaussian quadrature and Taylor series features. Compared to the popular Koopman theory and decomposition, which relies heavily on handcrafting a proper set of observables that linearize a nonlinear dynamics, our kernel operator-theoretic approach shifts the focus from feature engineering to adaptive filtering in a predefined RKHS. Simulation results show that this framework is robust and can outperform Koopman decomposition, ideally suited for applications where expert knowledge is absent and we must rely solely on the data.

In the future we will further reduce the computational complexity of the finite-dimensional explicit space FBF using dimensionality-reduction techniques and manifold learning. We will also apply this framework to model biological systems and signals, such as those observed in neuronal recordings in the brain, which are well-suited for data-driven model discovery techniques, and bio-inspired neuromorphic systems. These are ideally suited for leveraging modern data-driven modeling tools of machine learning to develop dynamical models characterizing the observed data.

\bibliographystyle{IEEEtran}
\bibliography{IEEEabrv,FbF_ref}{}

\begin{thebibliography}{10}
\providecommand{\url}[1]{#1}
\csname url@samestyle\endcsname
\providecommand{\newblock}{\relax}
\providecommand{\bibinfo}[2]{#2}
\providecommand{\BIBentrySTDinterwordspacing}{\spaceskip=0pt\relax}
\providecommand{\BIBentryALTinterwordstretchfactor}{4}
\providecommand{\BIBentryALTinterwordspacing}{\spaceskip=\fontdimen2\font plus
\BIBentryALTinterwordstretchfactor\fontdimen3\font minus
  \fontdimen4\font\relax}
\providecommand{\BIBforeignlanguage}[2]{{%
\expandafter\ifx\csname l@#1\endcsname\relax
\typeout{** WARNING: IEEEtran.bst: No hyphenation pattern has been}%
\typeout{** loaded for the language `#1'. Using the pattern for}%
\typeout{** the default language instead.}%
\else
\language=\csname l@#1\endcsname
\fi
#2}}
\providecommand{\BIBdecl}{\relax}
\BIBdecl

\bibitem{DL2015}
G.~H. Yann~LeCun, Yoshua~Bengio, ``Deep learning,'' \emph{Nature}, vol. 521,
  pp. 436--444, 2015.

\bibitem{SpeechDL2012}
G.~Hinton, L.~Deng, D.~Yu, G.~E. Dahl, A.-r. Mohamed, N.~Jaitly, A.~Senior,
  V.~Vanhoucke, P.~Nguyen, T.~N. Sainath, and B.~Kingsbury, ``Deep neural
  networks for acoustic modeling in speech recognition: The shared views of
  four research groups,'' \emph{IEEE Signal Processing Magazine}, vol.~29,
  no.~6, pp. 82--97, 2012.

\bibitem{ImageNetNIPS2012}
A.~Krizhevsky, I.~Sutskever, and G.~E. Hinton, ``Imagenet classification with
  deep convolutional neural networks,'' in \emph{Advances in Neural Information
  Processing Systems}, F.~Pereira, C.~Burges, L.~Bottou, and K.~Weinberger,
  Eds., vol.~25.\hskip 1em plus 0.5em minus 0.4em\relax Curran Associates,
  Inc., 2012.

\bibitem{GANsNIPS2014}
I.~Goodfellow, J.~Pouget-Abadie, M.~Mirza, B.~Xu, D.~Warde-Farley, S.~Ozair,
  A.~Courville, and Y.~Bengio, ``Generative adversarial nets,'' in
  \emph{Advances in Neural Information Processing Systems}, Z.~Ghahramani,
  M.~Welling, C.~Cortes, N.~Lawrence, and K.~Weinberger, Eds., vol.~27.\hskip
  1em plus 0.5em minus 0.4em\relax Curran Associates, Inc., 2014.

\bibitem{NLPsurvey2021}
D.~W. Otter, J.~R. Medina, and J.~K. Kalita, ``A survey of the usages of deep
  learning for natural language processing,'' \emph{IEEE Transactions on Neural
  Networks and Learning Systems}, vol.~32, no.~2, pp. 604--624, 2021.

\bibitem{interDL2017}
S.~Chakraborty, R.~Tomsett, R.~Raghavendra, D.~Harborne, M.~Alzantot,
  F.~Cerutti, M.~Srivastava, A.~Preece, S.~Julier, R.~M. Rao, T.~D. Kelley,
  D.~Braines, M.~Sensoy, C.~J. Willis, and P.~Gurram, ``Interpretability of
  deep learning models: A survey of results,'' in \emph{2017 IEEE
  SmartWorld/SCALCOM/UIC/ATC/CBDCom/IOP/SCI)}, 2017, pp. 1--6.

\bibitem{DLbrittle2019}
D.~Heaven, ``Why deep-learning ais are so easy to fool,'' \emph{Nature}, vol.
  574, no. 7777, pp. 163--166, 2019.

\bibitem{AIdynamicEnv2022}
A.~Iyer, K.~Grewal, A.~Velu, L.~O. Souza, J.~Forest, and S.~Ahmad, ``Avoiding
  catastrophe: Active dendrites enable multi-task learning in dynamic
  environments,'' \emph{Frontiers in Neurorobotics}, vol.~16, 2022.

\bibitem{Kalman60}
R.~E. Kalman, ``A new approach to linear filtering and prediction problems,''
  \emph{Trans. ASME, Series D., Journal of Basic Eng.}, vol.~82, pp. 35--45,
  1960.

\bibitem{Schmidt2009}
M.~Schmidt and H.~Lipson, ``Distilling free-form natural laws from experimental
  data,'' \emph{Science}, vol. 324, no. 5923, pp. 81--85, 2009.

\bibitem{RAISSI2019686}
M.~Raissi, P.~Perdikaris, and G.~Karniadakis, ``Physics-informed neural
  networks: A deep learning framework for solving forward and inverse problems
  involving nonlinear partial differential equations,'' \emph{Journal of
  Computational Physics}, vol. 378, pp. 686--707, 2019.

\bibitem{LEE2020108973}
K.~Lee and K.~T. Carlberg, ``Model reduction of dynamical systems on nonlinear
  manifolds using deep convolutional autoencoders,'' \emph{Journal of
  Computational Physics}, vol. 404, p. 108973, 2020.

\bibitem{ModernKoopman2022}
S.~L. Brunton, M.~Budi\v{s}i\'{c}, E.~Kaiser, and J.~N. Kutz, ``Modern koopman
  theory for dynamical systems,'' \emph{SIAM Review}, vol.~64, no.~2, pp.
  229--340, 2022.

\bibitem{Koopman31}
B.~O. Koopman, ``Hamiltonian systems and transformations in hilbert space,''
  \emph{Proceedings of the National Academy of Sciences of the United States of
  America}, vol.~17, no.~5, pp. 315--318, 1931.

\bibitem{Koopman32}
B.~O. Koopman and J.~V. Neumann, ``Dynamical systems of continuous spectra,''
  \emph{Proceedings of the National Academy of Sciences of the United States of
  America}, vol.~18, no.~3, pp. 255--263, 1932.

\bibitem{neumann1932pnas}
J.~v.~Neumann, ``Proof of the quasi-ergodic hypothesis,'' \emph{Proceedings of
  the National Academy of Sciences}, vol.~18, no.~1, pp. 70--82, 1932.

\bibitem{birkhoff1931pnas}
G.~D. Birkhoff, ``Proof of the ergodic theorem,'' \emph{Proceedings of the
  National Academy of Sciences}, vol.~17, no.~12, pp. 656--660, 1931.

\bibitem{rowley2009jfm}
C.~W. Rowley, I.~Mezi\'c, S.~Bagheri, P.~Schlatter, and D.~Henningson,
  ``Spectral analysis of nonlinear flows,'' \emph{Journal of Fluid Mechanics},
  vol. 641, p. 115–127, 2009.

\bibitem{mezic2005nd}
I.~Mezić, ``Spectral {Properties} of {Dynamical} {Systems}, {Model}
  {Reduction} and {Decompositions},'' \emph{Nonlinear Dynamics}, vol.~41,
  no.~1, pp. 309--325, Aug. 2005.

\bibitem{schmid2009dynamic}
P.~J. Schmid, ``Application of the dynamic mode decomposition to experimental
  data,'' \emph{Experiments in Fluids}, vol.~50, no.~4, pp. 1123--1130, Apr.
  2011.

\bibitem{brunton2019data}
S.~L. Brunton and J.~N. Kutz, \emph{Data-driven science and engineering:
  Machine learning, dynamical systems, and control}.\hskip 1em plus 0.5em minus
  0.4em\relax Cambridge University Press.

\bibitem{loeve1960probability}
M.~Lo{\`e}ve, \emph{Probability Theory}, ser. University series in higher
  mathematics.\hskip 1em plus 0.5em minus 0.4em\relax Van Nostrand, 1960.

\bibitem{Pearson1901}
K.~P. F.R.S., ``Liii. on lines and planes of closest fit to systems of points
  in space,'' \emph{Philosophical Magazine Series 1}, vol.~2, pp. 559--572,
  1901.

\bibitem{DMDvsOPD2015}
D.~A. Bistrian and I.~M. Navon, ``An improved algorithm for the shallow water
  equations model reduction: Dynamic mode decomposition vs pod,''
  \emph{International Journal for Numerical Methods in Fluids}, vol.~78, no.~9,
  pp. 552--580, 2015.

\bibitem{RKHS1950}
N.~Aronszajn, ``Theory of reproducing kernels,'' \emph{Transactions of the
  American Mathematical Society}, vol.~68, no.~3, pp. 337--404, 1950.

\bibitem{kernel2008}
T.~Hofmann, B.~Sch{\"o}lkopf, and A.~J. Smola, ``{Kernel methods in machine
  learning},'' \emph{The Annals of Statistics}, vol.~36, no.~3, pp. 1171 --
  1220, 2008.

\bibitem{kDMD2015}
``A kernel-based method for data-driven koopman spectral analysis,'' pp.
  247--265.

\bibitem{FUJII201994}
K.~Fujii and Y.~Kawahara, ``Dynamic mode decomposition in vector-valued
  reproducing kernel hilbert spaces for extracting dynamical structure among
  observables,'' \emph{Neural Networks}, vol. 117, pp. 94--103, 2019.

\bibitem{KAARMA}
K.~Li and J.~C. Pr{\'{\i}}ncipe, ``The kernel adaptive
  autoregressive-moving-average algorithm,'' \emph{IEEE Trans. Neural Netw.
  Learn. Syst.}, vol.~27, no.~2, pp. 334--346, Feb. 2016.

\bibitem{FBF2022}
K.~Li and J.~C. Príncipe, ``Functional bayesian filter,'' \emph{IEEE
  Transactions on Signal Processing}, vol.~70, pp. 57--71, 2022.

\bibitem{NoTrick2019}
K.~Li and J.~C. Pr{\'{\i}}ncipe, ``No-trick (treat) kernel adaptive filtering
  using deterministic features,'' \emph{CoRR}, vol. abs/1912.04530, 2019.

\bibitem{Bochner1959}
S.~Bochner, M.~Functions, S.~Integrals, H.~Analysis, M.~Tenenbaum, and
  H.~Pollard, \emph{Lectures on Fourier Integrals. (AM-42)}.\hskip 1em plus
  0.5em minus 0.4em\relax Princeton University Press, 1959.

\bibitem{Dao2017}
T.~Dao, C.~D. Sa, and C.~R{\'e}, ``Gaussian quadrature for kernel features,''
  in \emph{Proceedings of the 31st International Conference on Neural
  Information Processing Systems}, ser. NIPS'17.\hskip 1em plus 0.5em minus
  0.4em\relax USA: Curran Associates Inc., 2017, pp. 6109--6119.

\bibitem{UniversalKernels2006}
C.~A. Micchelli, Y.~Xu, and H.~Zhang, ``Universal kernels,'' \emph{Journal of
  Machine Learning Research}, vol.~7, no.~95, pp. 2651--2667, 2006.

\bibitem{Cotter2011}
A.~Cotter, J.~Keshet, and N.~Srebro, ``Explicit approximations of the gaussian
  kernel,'' \emph{CoRR}, vol. abs/1109.4603, 2011.

\bibitem{Zwicknagl2009}
B.~Zwicknagl, ``Power series kernels,'' \emph{Constructive Approximation},
  vol.~29, pp. 61--84, 2009.

\bibitem{Arasaratnam2009}
I.~Arasaratnam, ``Cubature {K}alman filtering: Theory \& applications,'' Ph.D.
  dissertation, McMaster University, 2009.

\bibitem{Mackey77}
M.~C. Mackey and L.~Glass, ``Oscillation and chaos in physiological control
  systems,'' \emph{Science}, vol. 197, no. 4300, pp. 287--289, Jul. 1977.

\bibitem{Ott02}
E.~Ott, \emph{Chaos in dynamical systems}, 2nd~ed.\hskip 1em plus 0.5em minus
  0.4em\relax Cambridge, UK: Cambridge University Press, 2002.

\bibitem{Haykin02}
S.~Haykin, \emph{Adaptive Filter Theory}, 4th~ed.\hskip 1em plus 0.5em minus
  0.4em\relax Prentice Hall, 2002.

\bibitem{Kutz2018}
J.~N. Kutz, J.~L. Proctor, and S.~L. Brunton, ``Applied koopman theory for
  partial differential equations and data-driven modeling of spatio-temporal
  systems,'' \emph{Complexity}, vol. 2018, pp. 1--16, 2018.

\end{thebibliography}

\appendix\label{Sec:Appendix}
Here, we show how to compute each of the submatrices in \eqref{Eq:SSM}.
Using the state transition \eqref{eq:exp_state_transition} and the Taylor series features \eqref{eq:taylor_features}, the state-transition Jacobian in \eqref{Eq:F1} becomes
\begin{align}
	\textbf{F}_1(i) & = \frac{\partial\textbf{A}_{i}\widetilde{\psi}(\textbf{s}_{i-1}) + \textbf{B}_{i} \widetilde{\phi}(\textbf{u}_{i})}{\partial \textbf{s}_{i-1}} \\
	& = \textbf{A}_i\frac{\partial \widetilde{\psi}(\textbf{s}_{i-1})}{\partial \textbf{s}_{i-1}}
\end{align}
where the $(l,m)$ element of $\textbf{F}_1(i)$ is
\begin{align}          
	\textbf{F}_1^{(lm)}(i)&=\frac{\partial \widetilde{\psi}(\textbf{s}_{i-1})^{(l)}}{\partial \textbf{s}_{i-1}^{(m)}}\\
	&=\frac{\partial e^{-\frac{\|\textbf{s}_{i-1}\|^2}{2\sigma^2}} \frac{1}{\sigma^{n}\sqrt{n!}}\binom{n}{\alpha^{(l)}}\textbf{s}_{i-1}^{\alpha^{(l)}}}{\partial \textbf{s}_{i-1}^{(m)}}\label{Eq:F1TS}
\end{align}                                                        where $\alpha^{(l)}=\left(\alpha_1^{(l)},\alpha_2^{(l)},\dots,\alpha_{n_s}^{(l)}\right)$ is the multi-indices notation for the multinomial expansion, $n = |\alpha^{(l)}|$, and $\textbf{s}^{\alpha^{(l)}}=\textbf{s}_1^{\alpha^{(l)}_1} \textbf{s}_2^{\alpha^{(l)}_2} \cdots \textbf{s}_{n_s}^{\alpha^{(l)}_{n_s}}$, with $m = 1,\cdots,n_s$ and $l = 1,\cdots,D$ denoting the $l^\text{th}$ feature or unique monomial of the D-dimensional TS approximation of the Gaussian kernel. We further expand \eqref{Eq:F1TS} using the product rule:
\begin{align}
	\textbf{F}(i)_{lm}&=\frac{e^{-\frac{\|\textbf{s}_{i-1}\|^2}{2\sigma^2}} \partial \frac{1}{\sigma^{n}\sqrt{n!}}\binom{n}{\alpha^{(l)}}\textbf{s}_{i-1}^{\alpha^{(l)}}}{\partial \textbf{s}_{i-1}^{(m)}} \\ 
	&\quad + \frac{\frac{1}{\sigma^{n}\sqrt{n!}}\binom{n}{\alpha^{(l)}}\textbf{s}_{i-1}^{\alpha^{(l)}} \partial e^{-\frac{\|\textbf{s}_{i-1}\|^2}{2\sigma^2}} }{\partial \textbf{s}_{i-1}^{(m)}}\\
	&= e^{-\frac{\|\textbf{s}_{i-1}\|^2}{2\sigma^2}} \frac{1}{\sigma^{n}\sqrt{n!}}\binom{n}{\alpha^{(l)}}\frac{\partial\textbf{s}_{i-1}^{\alpha^{(l)}}}{\partial \textbf{s}_{i-1}^{(m)}}\\
	&\quad + \frac{1}{\sigma^{n}\sqrt{n!}}\binom{n}{\alpha^{(l)}}\textbf{s}_{i-1}^{\alpha^{(l)}} e^{-\frac{\|\textbf{s}_{i-1}\|^2}{2\sigma^2}}\frac{\partial \frac{-\|\textbf{s}_{i-1}\|^2}{2\sigma^2}}{\partial \textbf{s}_{i-1}^{(m)}}\\
	&= e^{-\frac{\|\textbf{s}_{i-1}\|^2}{2\sigma^2}} \frac{1}{\sigma^{n}\sqrt{n!}}\binom{n}{\alpha^{(l)}}\frac{\alpha^{(l)}_m\textbf{s}_{i-1}^{\alpha^{(l)}}}{\textbf{s}_{i-1}^{(m)}}\\
	&\quad + \underbrace{\frac{1}{\sigma^{n}\sqrt{n!}}\binom{n}{\alpha^{(l)}}\textbf{s}_{i-1}^{\alpha^{(l)}} e^{-\frac{\|\textbf{s}_{i-1}\|^2}{2\sigma^2}}}_{\mathbbm{a}^{(l)}}\left(-\frac{\textbf{s}_{i-1}^{(m)}}{\sigma^2}\right)\\
	&=\mathbbm{a}^{(l)}\left(\frac{\alpha^{(l)}_m}{\textbf{s}_{i-1}^{(m)}}-\frac{\textbf{s}_{i-1}^{(m)}}{\sigma^2}\right).
\end{align}

We can compute \eqref{Eq:F2A} and \eqref{Eq:F2B} directly as
\begin{align}
	\textbf{F}^{(\textbf{A})}_2(i) & = \frac{\partial \textbf{s}_i}{\partial \textbf{A}(i)}\nonumber\\
	& = \operatorname{diag}(\widetilde{\psi}(\textbf{s}_{i-1})^\intercal, \dots, \widetilde{\psi}(\textbf{s}_{i-1})^\intercal) \\
	&\quad \in \mathbb{R}^{n_s\times\left(n_s n_{\widetilde{\psi}(s)}\right)}\nonumber\\
	\intertext{and} 
	\textbf{F}^{(\textbf{B})}_2(i) & = \frac{\partial \textbf{s}_i}{\partial \textbf{B}(i)}\nonumber\\
	& = \operatorname{diag}(\widetilde{\phi}(\textbf{u}_{i})^\intercal, \dots, \widetilde{\phi}(\textbf{u}_{i})^\intercal) \\
	&\quad \in \mathbb{R}^{n_s\times\left(n_s n_{\widetilde{\phi}(u)}\right)}.\nonumber
\end{align}                                
However, a more efficient organization is to break the weight matrix vector $\bm{\Omega}_i$ in (\ref{Eq:AugState}) into individual state dimension components and rewrite (\ref{Eq:SSM}) as
\begin{align}
	\begin{bmatrix}
		\textbf{s}_i \\[0.3em]
		\bm{\Omega}^{(k)}_i\\[0.3em]
	\end{bmatrix} = \begin{bmatrix}
		\textbf{F}_1 & \textbf{F}^{(k)}_2 \\[0.3em]
		\textbf{0} & \textbf{I}_{n_{\bm{\Omega}^{(k)}}} \\[0.3em]
	\end{bmatrix}\begin{bmatrix}
		\textbf{s}_{i-1} \\[0.3em]
		\bm{\Omega}^{(k)}_i\\[0.3em]                           
	\end{bmatrix} + \bm{w}^{(k)}_{i-1} \label{Eq:F2_perState}
\end{align}
where $\bm{\Omega}^{(k)}_i$ are all the weights connected to the $k$-th output state node ($1\leq k\leq n_s$) and
\begin{align}                                                                                          
	\textbf{F}_2^{(k)}(i) \stackrel{\Delta}{=}\frac{\partial \textbf{s}_i}{\partial \bm{\Omega}^{(k)}_i} = [\widetilde{\psi}(\textbf{s}_{i-1})^\intercal, \widetilde{\phi}(\textbf{u}_{i})^\intercal] \label{Eq:F2new}.
\end{align}
At each time step $i$, this process is repeated for each of the $n_s$ state components.

For the case when $\textbf{s}$ is observable (e.g., known nonlinear dynamics) as shown in \eqref{eq:RKHS_state_transition}, we can define the state transitions directly in the RKHS, i.e., 
\begin{align}
	\begin{bmatrix}
		\widetilde{\psi}(\textbf{s}_{i}) \\[0.3em]
		\bm{\Omega}^{(k)}_i\\[0.3em]
	\end{bmatrix} = \begin{bmatrix}
		\textbf{F}_1 & \textbf{F}^{(k)}_2 \\[0.3em]
		\textbf{0} & \textbf{I}_{n_{\bm{\Omega}^{(k)}}} \\[0.3em]
	\end{bmatrix}\begin{bmatrix}
		\widetilde{\psi}(\textbf{s}_{i-1}) \\[0.3em]
		\bm{\Omega}^{(k)}_i\\[0.3em]                         
	\end{bmatrix} + \bm{w}^{(k)}_{i-1} \label{Eq:F1_linear}
\end{align}
where corresponding state-transition matrix simplify to
\begin{align}
	\textbf{F}_1(i) & = \frac{\partial\textbf{A}_{i}\widetilde{\psi}(\textbf{s}_{i-1}) + \textbf{B}_{i} \widetilde{\phi}(\textbf{u}_{i})}{\partial \widetilde{\psi}(\textbf{s}_{i-1})} \\
	& = \textbf{A}_i.
\end{align}

\begin{IEEEbiography}
	[{\includegraphics[width=1in,height=1.25in,clip,keepaspectratio]{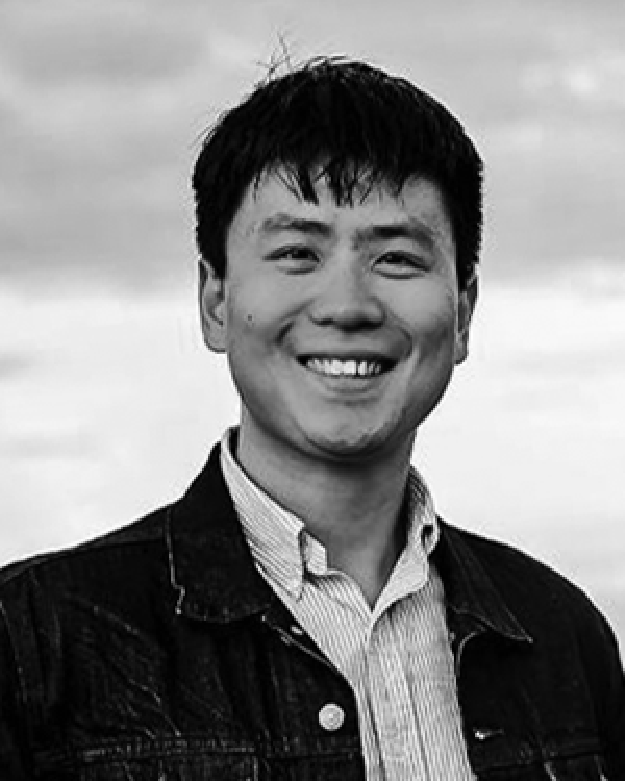}}]
	{Kan Li} (S'08) received the B.A.Sc. degree in electrical engineering from the University of Toronto in 2007, the M.S. degree in electrical engineering from the University of Hawaii in 2010, and the Ph.D. degree in electrical engineering from the University of Florida in 2015.  He is currently a research scientist at the University of Florida. His research interests include machine learning and signal processing.
\end{IEEEbiography}
\vskip -2\baselineskip plus -1fil
\begin{IEEEbiography}
	[{\includegraphics[width=1in,height=1.25in,clip,keepaspectratio]{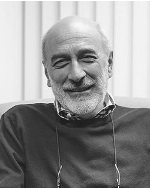}}]{Jos\'{e} C. Pr\'{i}ncipe}
	(M'83-SM'90-F'00) is the BellSouth and Distinguished Professor of Electrical and Biomedical Engineering at the University of Florida, and the Founding Director of the Computational NeuroEngineering Laboratory (CNEL). His primary research interests are in advanced signal processing with information theoretic criteria and adaptive models in reproducing kernel Hilbert spaces (RKHS), with application to brain-machine interfaces (BMIs). Dr.  Pr\'{i}ncipe is a Fellow of the IEEE, ABME, and AIBME. He is the former Editor in Chief of the IEEE Transactions on Biomedical Engineering, past Chair of the
	Technical Committee on Neural Networks of the IEEE Signal Processing Society, Past-President of the International Neural Network Society, and a recipient of the IEEE EMBS Career Award, the IEEE Neural Network Pioneer Award, and the IEEE SPS Claude Shannon-Harry Nyquist Technical Achievement Award.
\end{IEEEbiography}	
\end{document}